\documentclass{article}

\usepackage[numbers]{natbib}

\usepackage[final]{neurips_2024}

\usepackage[utf8]{inputenc} 
\usepackage[T1]{fontenc}    
\usepackage{hyperref}       
\usepackage{url}            
\usepackage{booktabs}       
\usepackage{amsfonts}       
\usepackage{nicefrac}       
\usepackage{microtype}      
\usepackage{xcolor}         
\usepackage{csquotes}

\usepackage{graphicx}
\usepackage{booktabs}

\usepackage[accsupp]{axessibility}  
\usepackage{url}            
\usepackage{booktabs}       
\usepackage{amsfonts}       
\usepackage{nicefrac}       
\usepackage{microtype}      
\usepackage{xcolor}         
\usepackage{subcaption}
\usepackage[ruled]{algorithm2e}
\SetAlFnt{\small}
\usepackage{amsmath}
\usepackage{bm}
\usepackage{bbold}
\usepackage{graphicx}
\usepackage{multirow}
\usepackage{textcomp}
\usepackage{wrapfig}
\usepackage{subcaption}
\usepackage{tabularx}
\usepackage{xspace}
\usepackage{interval}
\usepackage{bm,upgreek}
\usepackage{colortbl}
\usepackage{enumitem}
\usepackage{amssymb}
\usepackage{pifont}
\usepackage{wrapfig,lipsum,booktabs}
\usepackage{tabularx}
\usepackage{booktabs}
\usepackage{adjustbox}
\usepackage{ dsfont }
\usepackage{epigraph}
\usepackage{amsthm}
\newcommand{\cmark}{\ding{51}} 
\newcommand{\xmark}{\ding{55}} 

\usepackage{titlesec}

\titlespacing*{\section}{0pt}{1pt}{1pt}
\titlespacing*{\subsection}{0pt}{1pt}{1pt}

\title{

Asynchronous Perception Machine \\
for Test Time Training
}
\author{%
  Rajat Modi\thanks{Correspondence to \href{mailto:rajatmodi62@gmail.com}{\texttt{rajatmodi62@gmail.com}}.
}  \quad  Yogesh Singh Rawat\\
  Centre for Research in Computer Vision\\
  University of Central Florida\\
  Orlando, FL 32765 \\
  \texttt{rajatmodi62@gmail.com,yogesh@crcv.ucf.edu} \\
  \href{https://rajatmodi62.github.io/apm_project_page}{https://rajatmodi62.github.io/apm\_project\_page}
}

\begin{document}

\maketitle

\vspace{-2em}
\begin{abstract}

In this work, we propose Asynchronous Perception Machine (APM), a computationally-efficient architecture for test-time-training (TTT). APM can process patches of an image one at a time in any order \textit{asymmetrically,} and \textit{still encode} semantic-awareness in the net. We demonstrate APM's ability to recognize out-of-distribution images \textit{without} dataset-specific pre-training, augmentation or any-pretext task. APM offers competitive performance over existing TTT approaches. To perform TTT, APM just distills test sample's representation \textit{once}. APM possesses a unique property: it can learn using just this single representation and starts predicting semantically-aware features. 

APM demostrates potential applications beyond test-time-training: APM can scale up to a dataset of 2D images and yield semantic-clusterings in a single forward pass. APM also provides first empirical evidence towards validating GLOM's insight, i.e. if input percept is a field. Therefore, APM helps us converge towards an implementation which can  do \textit{both} interpolation and perception on a \textit{shared}-connectionist hardware. 

\vspace{-1em}
\end{abstract}


\section{Introduction}

In these past centuries, computing-machines have become a lot faster \cite{turing1936computable}. This made it possible to train higher-parameterized neural nets and led to interesting emergent abilities 
 \cite{schaeffer2024emergent}. As was predicted by Turing himself, and as were his suspicions of Lady Lovelace's arguments \textit{against} learning machines\cite{turing2009computing}: neural-nets can now finally learn without human-feedback \cite{caron2021emerging}, paint pictures \cite{rombach2022high} and even compose a sonnet \cite{turing1936computable}. Even with such impressive-progress, a key question still remains: how can these nets recognize images 
whose distribution is far different from the ones which were used during training? For e.g., consider a self-driving car trying to stop when it encounters a pedestrian crossing a road. Such practical scenarios require `instantaneous-decisions' for ensuring human-safety in autonomous-systems \cite{modi2023occlusions}. 


Test-time-training (TTT) \cite{sun2020test} is  one of the promising techniques for handling such distribution shifts: a neural net adapts to a test sample `on the fly'. Since the label of the sample is not known, the net performs some auxiliary pre-text task like data augmentation \cite{ttt_mae}, rotation \cite{ttt_mae,ttt_sun} or prompt tuning \cite{tpt} on it. After several such iterations, the net recognizes the test sample. The key idea is that the net is allowed to \textit{dynamically} adjust its decision boundary even \textit{after} it has been trained, thereby bringing it much closer to how humans keep learning `continuously' throughout their lifespans \cite{ttt_sun}.

Despite the success of existing TTT approaches, several limitations need to be addressed \cite{tpt,ttt_mae,ttt_sun}:
1) \textbf{The Information Bottleneck Problem} \cite{modi2023occlusions}: Multiple TTT iterations requires feed-forward through many hidden layers \textit{multiple times}, making it computationally expensive. 2) \textbf{Reliance on a surrogate pre-text task:} the optimal data-augmentation pipeline or the best pretext task is not known beforehand, worsening the issue even further in an online setting. 3) Furthermore, TTT leverages architectures like transformers which rely on \textbf{parallel perception}:  this requires projecting all input patches into a shared representational space, thereby consuming significant memory.

\textbf{Inspired by GLOM's philosophy} \cite{hinton2023represent}, we hereby propose Asynchronous Perception Machine (APM), a new architecture for efficient test-time-training. 1) It handles the information-bottleneck problem by directly learning a shortcut from input image to final representation from last layer of a model \cite{hinton2022forward}. 2) During TTT, we compute test-sample's representation \textit{only once}. Subsequent iterations involve over-fitting on this representation only and \textit{doesn't} require any data-augmentation/pretext task. 3) APM can operate on a \textit{single} patch at any location \textit{asynchronously} \cite{turing1990chemical} and still encode semantic-awareness in the net, thereby offering a \textit{fresh perspective} towards machine-perception.

We make the following contributions in this work:

\vspace{-\baselineskip}
\begin{itemize}[noitemsep]
\item We propose APM, a GLOM-inspired architecture that can perform test-time-training \textit{without} requiring data augmentation/auxilary-tasks. APM is a step towards validating GLOM's philosophical insight: a percept is \textit{really} a field \cite{hinton2023represent}. 
\item APM is computationally efficient, i.e it almost \textit{halves} the number of FLOPs over existing methods \cite{ttt_mae,ttt_sun}. APM matches/surpasses TTT performance by $0.4\%-8\%$ across $16$ datasets.
\item APM is architecturally simple: a convolutional layer and a single MLP of 5 layers. APM can still learn using one sample and semantically clusters on a collection of images \cite{hinton2022forward}.  
\end{itemize}

\begin{figure}[t]
\vspace{-2em}
    \centering
    
    \includegraphics[width=\linewidth]{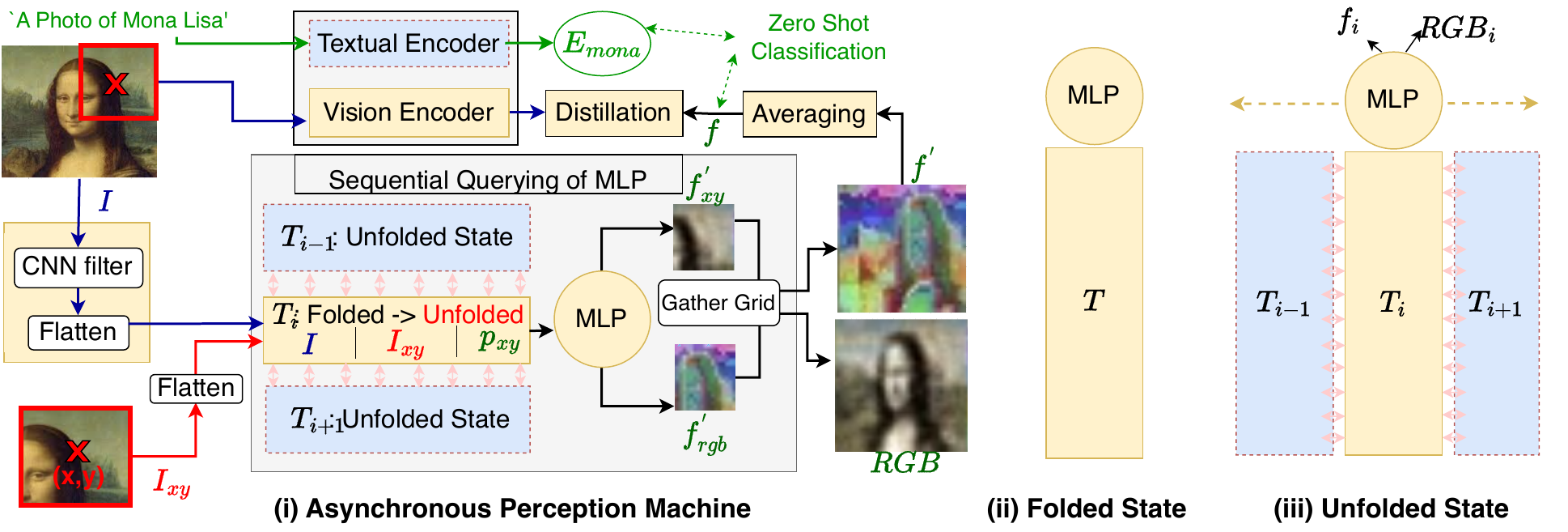}
    
       \caption{\textbf{(i) Asynchronous Perception Machine (APM)}:   An image $I$ passes through a column module and routes to a trigger column $T_{i}$. $T_{i}$ then unfolds and generates $h \times w$ location-specific queries. These queries are i.i.d and can be parallelized across cores of a gpu \cite{krizhevsky2014one}. Each query $T_{i}$ is passed through a \textit{shared} MLP and yields the vector $f'_{xy}$ and $f'_{rgb}$. MLP is queried iteratively until whole grid $f^{'}$ comes into existence. Classification then involves comparing the averaged representation $f$ with class-specific textual representations in the contrastive space . \textbf{(ii) Folded State}: The parameters which the net learns are parameters of $T$ and MLP. \textbf{(iii) Unfolded State}: $T$ expands to yield $h \times w$ queries `on-the-fly'. Learning involves oscillating this net between folded and unfolded states. This net can be trained via backpropogation \cite{rumelhart1986learning,hinton2022forward}.       
       }

    \label{fig:architecture}
\vspace{-1em}
\end{figure}

\section{Background}

We draw from insights previously philosophically mentioned in GLOM \cite{hinton2023represent}. Consider how machine perception has been done classically: an input $x \in \mathds R^{c \times h\times w}$ is transformed by a non-linear function $f$ to a perceptual feature-cuboid $y \in \mathds R^{ c' \times l \times h\times w}$, $c$ denotes image channels, $l$ are the intermediate layers, $h,w$ are input spatial dimensions. Note that $y$ is estimated after feed-forward through $l$ layers. 


 Next, we imagine feed-forwarding  one patch $(x,y)$ into the neural net $f$ and estimating the response $v_{c'}$. It thus becomes possible to process these patches one-by-one rather than keeping them all together in memory \cite{vaswani2017attention}. We then further imagine auto-regressively querying $f$ until the whole grid $y$ has been brought into existence.

\textbf{Implicit-representations/Neural-Fields:} The neural net $f$  inputs a coordinate based query, $(I,x,y)$ where $(x,y)$ is the 2D location in an image $I$. It then gives the answer $v_{c'}$ \cite{hinton2023represent}. Such implicit representations have been studied extensively in 3D novel-view synthesis \cite{mildenhall2021nerf}. However, in the above problem $I$ is a simple 2D image. 
In a recent work \cite{hinton2023represent}, it was hypothesized that neural-implicit representations can work on 2D images as well without having to retrain it every-time. 

\textbf{Layer Skipping \cite{hinton2022forward}}: We now combine the previous two insights together. Imagine that $v_{c'}$ has been estimated from a model after travelling through several layers. We can distill these $v_{c'}$ into an implicit representation $f$ and learn a direct mapping between input $x$ and last layer of a model \cite{hinton2022forward}. During inference, we can skip feed-forwarding through \textit{all} the model layers. The only feed-forward cost we then pay is what it costs to travel through $f$. Provided that the number of parameters in $f$ are lesser than parameters in the teacher model, \textit{we can expect computational speedups.}

We now introduce a connectionist-net $f$ motivated by these insights. It can perform efficient test-time-training on a given 2D input.

\section{Asynchronous Perception Machine (APM)}
\label{sec:architecture}

APM processes an input image $I$ via two novel mechanisms: 1) A column module $T$ which is said to contain an input image $I$, 2) a column folding-unfolding mechanism that operates during each forward-pass. We first provide a technical analogy to better understand APM.  

\subsection{A Technical Analogy\cite{hinton2023represent}}
A neural field does 3D novel view synthesis by querying an MLP with a point $(x,y,z)$ and yielding corresponding rgb. In a similar way, APM does \textit{2D perception} by querying an MLP with an image $I$, and a location $(x,y)$ to yield location-specific feature $f_{xy}$. APM features a new mechanism to query the MLP, i.e \textit{a column module} $T$. Next, we define this column representation $T$. 

\subsection{Global Column Module: Defining compressed representation T}
\label{sec:trigger}

We define a column $T$ as a vector of dimensions $1 \times 1 \times d$. Our aim is to map image $I$ in this $T$, so that $T$ can summarize its entire identity. Given an image of dimensions $c \times h \times w$, we run a 2D convolution on it. Number of filters are set as accordingly. The resultant  $1\times h' \times w'$ feature map is then flattened into a single column vector T of dimensions $d = h' \times w'$ . We shall refer to this column T as "triggering hyper-column"(seed). \emph{The only learnable parameters in this column are parameters of a convolution filter.}\footnote{The word triggering has been chosen because this column shall be used to trigger the queries to the neural field shared across different locations \cite{hinton2023represent}.}\cite{turing1936computable,turing1990chemical}.

\subsection{Abstract view: The Column Unfolding-Folding Mechanism}
\label{sec:column_mechanism}

The trigger column T now starts undergoing cycles of folding-unfolding (Fig \ref{fig:architecture}). During unfolding, the column T copies itself to yield $h \times w$ \textit{location-aware} columns. During folding, these $ h \times w$ columns collapse back into a single column $T$. The neural-net then \textit{oscillates} between these folded-unfolded phases during learning iterations.

\subsection{Computational Principle: Location-aware columns and their collapse}
\label{sec:query_existence}
\textbf{The unfolding-process} shall now be concerned with generating location-aware column $T_{ij}$ from $T$. We generate $h*w$ 2-D \textit{non-parametric} positional encoding similar to the ones being used in transformers \cite{vaswani2017attention} and neural fields \cite{mildenhall2021nerf}. The trigger column $T_{ij}$ is then given by $T_{ij} = (T| p_{ij})$, where $|$ denotes the stacking operator and $(1,1) \leq (i,j) \leq (H,W)$. T can be said to encode \textit{identity} of an image.

\textbf{The folding-process} involves collapsing all columns $T_{ij}$ back into $T$ from which they had begun. This is achievable since the $p_{ij}$ used in $T_{ij}$ was deterministic. An abstract-mathematical intuition on folding is that \textit{all} $p_{ij}$ in $T_{ij}$ get deleted/annihilated at the end of every forward-pass, and only $T$ is left. Positional-encodings contain periodic-sinusoids which offers a strong positional-prior in practice \cite{vaswani2017attention}. Gradients collected from all $T_{ij}$ then update the parameters of the convolutional filter in the column $T$. This sharing of T across different locations $(i,j)$ now induces a new fundamental behavior \cite{hinton2023represent} already hypothesized in GLOM (more details in  section \ref{sec:continous_interpolation}).  

 This representation $T_{ij} = (T| p_{ij})$ exhibits a strong-symmetry breaking behavior \cite{hinton2023represent}\footnote{It means generating a unique representation $T_{ij}$ for any image and a location selected in it. Another way to break symmetry is to add noise \cite{hinton1984boltzmann}. Noise also helps escape degenerate local-minima. These ideas can be traced back to boltzmann-machines/hopfield-nets\cite{hopfield1982neural}. }. For e.g., consider different locations across \textit{same} image $I$. Although the identity T will be the same, the positional encoding $p_{ij}$ shall \textit{differentiate} among different locations. The converse holds true: given \textit{the same} position $p_{ij}$ in two \textit{different} images, the column $T$ shall \textit{differentiate} among identities of different images. 

 All the experiments in this paper are performed \textit{without-injecting} the local patch $I_{xy}$ in the trigger column $T_{ij}$. This showcases the strong-symmetry breaking behaviour of positional-encodings\cite{vaswani2017attention}: they can disentangle location-information from \textit{global} $I$ \textit{without-requiring} an additional patch-prior. 

\subsection{Firing location-aware columns into a shared MLP}
\label{sec:col_firing}
Each column $T_{ij}$ is passed through an MLP to yield location-specific features $f_{ij}$ and RGB values $RGB_{ij}$. Number of neurons in the first layer of the MLP is same as dimensionality of column $T_{ij}$. 

\textbf{Column Independence:} The MLP is shared across all the columns. One column is also independent of another as illustrated in Fig\ref{fig:architecture}. Therefore, the MLP can be queried \textit{sequentially}.  Firing a column $T_{ij}$  into the MLP yields a column vector $v_c$. Once $h \times w$ columns are finished firing, we get a feature grid $f$ of dimensions $h \times w \times c$. Note that the number of columns can be as low as 1.

\subsection{Training and Losses}

We detail how the $t^{th}$ iteration of test-time-training could be performed. First, the obtained feature grid $f \epsilon \mathds{R}
^{h \times w \times c}$ from APM is averaged to yield $f_{avg} \epsilon \mathds{R}
^{c}$. For the first TTT iteration, i.e. $t=1$, the image $I$ is feed-forwarded through a multi-modal teacher like CLIP to get a CLS token $f_{cls}$ and corresponding text representation $t_{cls}$. APM then learns to estimate this \textit{same} target feature $f_{cls}$. We enforce this by a simple $L_2$ constraint as:
\begin{equation}
L_{i} =  L_2(f_{avg,t}, f_{cls}) 
\end{equation}
where $f_{avg,t}$ is the averaged output feature from APM at a particular TTT iteration t. Note that \textbf{the target $f_{cls}$ is only estimated in $t=1$ and remains the same for $t \geq 2$,} i.e. subsequent feed-forward through teacher is \textit{not} needed. 

\textbf{Memory-efficient estimation of $f_{avg,t}$}: During a TTT iteration $t$, $f_{avg}$ is computed as a simple average of $f \epsilon \mathds{R}
^{h \times w \times c}$. This would require $h \times w$ columns to exist in the memory \textit{simultaneously}. APM's design assumes column independence which allows estimating $f_{avg}$ as a statistical running average, i.e. 
\begin{equation}
f_{avg} = \frac{n\times f_{avg} +  f_{i,j}}{n+1}
\end{equation}
assuming, $n$ columns have already been fired into the APM and one additional column corresponding to position $(i,j)$ of image $I$ is in the process of firing. This procedure is repeated until all positions $(i,j)$ are exhausted. We represent the sequential column-firing by a `Gather-Grid' operator in Fig\ref{fig:architecture}.

\textbf{Predicting image class-label:} After certain TTT iterations, (say t = 20), the output feature $f_{avg,t}$ and the textual features $t_{cls}$ are obtained. Image-classification then follows the standard practice of comparing the distance of $f_{avg,t}$ with each plausible class feature $t_{cls}$ in the contrastive space and choosing the closest one as the prediction \cite{radford2021learning}. 
\section{Experimenting with APM}
We quantitatively evaluate APM on \textit{zero-shot} test-time-training on popular benchmarks containing distribution-shifts \cite{ttt_mae,ttt_sun,tpt}. Next, we quantitatively explore its computational efficiency. 

\textbf{Datasets:} Cifar-10C \cite{ttt_sun} contains $5$ level of corruptions on the test-set with $15$ types of noises. \textit{Larger} datasets with significant distribution shifts consists of ImageNet val/other curated splits. For e.g., \textbf{ImageNet-V2} contains natural images consisting of $10k$ images across $1000$ classes. \textbf{ImageNet-A} contains $7500$ adversarial-images consisting of $200$ categories. \textbf{ImageNet-R} consists of $30000$ artistic-images across $200$ ImageNet categories. \textbf{ImageNet-Sketch} consists of black/white sketches of $50000$ images for $1000$ classes. \textbf{ImageNet-C} consisting of $15$ types of corruptions with $5$ levels of severity. There are additional $9$ cross-dataset generalization datasets \cite{tpt}.

\textbf{Baselines:} We compare against standard TTT-online \cite{ttt_sun}, TTT-MAE \cite{ttt_mae}, TPT \cite{tpt}, CLIP VIT-B/16, Coop, CocoOP. We also benchmark CLIP VIT-L/14 and the strongest OpenCLIP VIT-H/14-quickgelu variant pre-trained on \textit{dfn5b}.

 %
\begin{table}[ht]
  \vspace{-1em}
  \caption{\textbf{APM's Robustness to Natural Distribution Shifts}. CoOp and CoCoOp are tuned on ImageNet using 16-shot training data per category. Baseline CLIP, prompt ensemble, TPT and our APM do not require training data. A \ding{51} in P means that method leveraged \textbf{pre-trained weights} on clean variant of train set aka, Image-net and downstream-ttt on corrupted version.}
  \label{tab:imagenet_splits}  
  \centering    
  \resizebox{\linewidth}{!}{%
  \begin{tabular}{l*{8}{c}}
    \toprule
    \multirow{2}{*}{Method} & \multirow{2}{*}{P} & ImageNet & ImageNet-A & ImageNet-V2 & ImageNet-R & ImageNet-Sketch & \multirow{2}{*}{Average} & \multirow{2}{*}{OOD Average} \\
                            &                          & Top1 acc. $\uparrow$ & Top1 acc. $\uparrow$ & Top1 acc. $\uparrow$ & Top1 acc. $\uparrow$ & Top1 acc. $\uparrow$ & & \\
    \midrule
   CLIP-ViT-B/16           & \ding{55}                & 66.7 & 47.8 & 60.8 & 73.9 & 46.0 & 59.1 & 57.2 \\
    Ensemble                & \ding{55}                & 68.3 & 49.8 & 61.8 & \textbf{77.6} & 48.2 & 61.2 & 59.4 \\
    
    TPT                     & \ding{55}                & \textbf{68.9} & \textbf{54.7} & 63.4 & 77.0 & 47.9 & 62.4 & 60.8 \\
    APM (Ours)              & \ding{55}                & 68.1 & 52.1 & \textbf{67.2} & 76.5 & \textbf{49.3} & \textbf{62.6} & \textbf{61.2} \\
    \hline
    \rowcolor[HTML]{F5F5F5} \textcolor{gray}{CoOp} & \ding{51} & 71.5 & 49.7 & 64.2 & 75.2 & 47.9 & 61.7 & 59.2 \\
    \rowcolor[HTML]{F5F5F5} \textcolor{gray}{CoCoOp} & \ding{51} & 71.0 & 50.6 & 64.0 & 76.1 & 48.7 & 62.1 & 59.9 \\
    \rowcolor[HTML]{F5F5F5} \textcolor{gray}{TPT + CoOp} & \ding{51} & 73.6 & 57.9 & 66.8 & 77.2 & 49.2 & 64.9 & 62.8 \\
    \rowcolor[HTML]{F5F5F5} \textcolor{gray}{TPT + CoCoOp} & \ding{51} & 71.0 & 58.4 & 64.8 & 78.6 & 48.4 & 64.3 & 62.6 \\
    \hline
    
    CLIP VIT-L/14           & \ding{55}                & 76.2 & 69.6 & 72.1 & 85.9 & 58.8 & 72.5 & 71.6 \\
    APM (Ours)              & \ding{55}                & \textbf{77.3} & \textbf{71.8} & \textbf{72.8} & \textbf{87.1} & \textbf{62.2} & \textbf{74.2} & \textbf{73.4} \\
    \midrule
    
    OpenCLIP-VIT-H/14       & \ding{55}                & 81.6 & 79.1 & 80.7 & 92.9 & 72.8 & 81.4 & 81.3 \\
    APM (Ours)              & \ding{55}                & \textbf{84.6} & \textbf{84.2} & \textbf{83.9} & \textbf{94.9} & \textbf{77.1} & \textbf{84.9} & \textbf{85.0} \\
    \hline
  \end{tabular}
}
\vspace{-1em}
\end{table}

\begin{table}[ht]
\caption{\textbf{APM's performance on ImageNet-C, level 5}. The first three rows are fixed models without test-time training. The third row, ViT probing, is the baseline used in \cite{ttt_mae}. A \ding{51} in P means that method leveraged \textbf{pre-trained weights} on clean variant of train set aka, Image-net and downstream-ttt on corrupted version. CLIP VIT-L/14 is generally more robust. APM does better on $11/15$ noises with an average accuracy score of $50.3$.}
\label{tab:imagenet_c}
\centering    
\resizebox{\linewidth}{!}{%
\begin{tabular}{l*{18}{c}}
\toprule
& P & brigh & cont & defoc & elast & fog & frost & gauss & glass & impul & jpeg & motn & pixel & shot & snow & zoom & Average \\
\midrule
\rowcolor[HTML]{EFEFEF}
Joint Train & \ding{51} & 62.3 & 4.5 & 26.7 & 39.9 & 25.7 & 30.0 & 5.8 & 16.3 & 5.8 & 45.3 & 30.9 & 45.9 & 7.1 & 25.1 & 31.8 & 24.8 \\
\rowcolor[HTML]{EFEFEF}
Fine-Tune & \ding{51} & 67.5 & 7.8 & 33.9 & 32.4 & 36.4 & 38.2 & 22.0 & 15.7 & 23.9 & 51.2 & 37.4 & 51.9 & 23.7 & 37.6 & 37.1 & 33.7 \\
\rowcolor[HTML]{EFEFEF}
ViT Probe & \ding{51} & 68.3 & 6.4 & 24.2 & 31.6 & 38.6 & 38.4 & 17.4 & 18.4 & 18.2 & 51.2 & 32.2 & 49.7 & 18.2 & 35.9 & 32.2 & 29.2 \\
\rowcolor[HTML]{EFEFEF}
TTT-MAE & \ding{51} & 69.1 & 9.8 & 34.4 & 50.7 & 44.7 & 50.7 & 30.5 & 36.9 & 32.4 & 63.0 & 41.9 & 63.0 & 33.0 & 42.8 & 45.9 & 44.4 \\
\midrule 
OpenCLIP VIT-L/14 & \ding{55} & 71.9 & 47.0 & 50.3  & 32.7 & 58.3 & 46.9 & 26.0 & 26.5 & 28.1 & 62.7 & 37.7 & 58.3 & 28.2 & 50.4 & 37.9 & 42.1 \\
APM (Ours) & \ding{55} & \textbf{77.4} & \textbf{51.9} & \textbf{56.6} & \textbf{37.9} & \textbf{64.8} & \textbf{53.2} & \textbf{28.7} & \textbf{31.4} & \textbf{33.0} & \textbf{68.4} & \textbf{44.1} & \textbf{64.5} & \textbf{33.1} & \textbf{56.9} & \textbf{43.9} & \textbf{50.3} \\
\bottomrule
\end{tabular}
}
\end{table}


\textbf{Results and Analysis:} We study APM's performance on test-time-training on several datasets. APM processes each test sample individually: i.e. the weights are drawn from a normal distribution after processing every sample to prevent information leakage. For zero-shot classification of a test-sample, APM leverages the $80$ textual-prompt templates similar to the ones used in CLIP. In Tab \ref{tab:imagenet_splits}, APM scales up to \textit{zero-shot} classification task to datasets with $1000$ classes. Using CLIP-ViT B/16 as a teacher, we surpass TPT \cite{tpt} with an avg score of $62.6$ and avg ood-score of $61.2$. Next, we benchmark OpenCLIP-VITH/14 against all these splits. Using the same model as our teacher, we get an absolute improvement of $3\% \uparrow$ $[84.6\%]$ on ImageNet val set, $5.1\% \uparrow $ $[84.2\%]$ on ImageNet-A, $3.2\% \uparrow $ $[83.9\%]$ on ImageNet-V2, $2\% \uparrow $ $[94.9\%]$ on ImageNet-R, $4.3\% \uparrow$ $[77.1\%]$ on ImageNet-Sketch respectively. A similar trend is observed with a VIT-L/14 backbone. This \textit{might} lead to the conclusion that a stronger teacher seems to benefit APM. In Table \ref{tab:imagenet_c}, we show results on Imagenet-C. Using a CLIP VIT-L/14 teacher, APM gets highest accuracy on $11/15$ noises with the highest average score of $50.3$. Note that TTT-MAE also uses a VIT-L encoder, and is $pre\-trained$. In contrast, $APM$ \textit{doesn't} need any dataset specific-pretraining.  Finally, in Tab \ref{tab:fine-grained}, APM improves upon $4/9$ datasets, comes close on remaining $5$ and gets the highest average accuracy score of $65.5$.

\begin{table}[ht]
\vspace{-2em}
  \caption{\textbf{Cross-dataset generalization} from ImageNet to fine-grained classification datasets. CoOp and CoCoOp are tuned on ImageNet using 16-shot training data per category. Baseline CLIP, prompt ensemble, TPT and APM do not require training data or annotations. We report top-1 accuracy.}
 
  \centering
  \resizebox{\linewidth}{!}{%
 
  \begin{tabular}{l*{11}c}
    \toprule
     \multirow{2}{*}{Method}      & \multirow{2}{*}{P}    & \multirow{2}{*}{Flower102}   & \multirow{2}{*}{DTD}  & \multirow{2}{*}{Pets}  & \multirow{2}{*}{UCF101}   & \multirow{2}{*}{Caltech101}  & \multirow{2}{*}{Food101} & \multirow{2}{*}{SUN397} & \multirow{2}{*}{Aircraft} & \multirow{2}{*}{EuroSAT} & \multirow{2}{*}{Average} \\
     &   &    &    &    &    &    &    &    &    &   &  \\
  \midrule
  \rowcolor[HTML]{F5F5F5} \textcolor{gray}{CoOp}            & \ding{51} &  68.7   & 41.9  &  89.1    &  66.5   &   93.7   &  85.3   &   64.2    &   18.5    &   46.4     &    63.9      \\

  \rowcolor[HTML]{F5F5F5} \textcolor{gray}{CoCoOp} & \ding{51} & 70.9  & 45.5  &  90.5    & 68.4   &   93.8   &  84.0   &   66.9    &   22.3    &   39.2     &    64.6      \\
  \hline
  
  CLIP-ViT-B/16   & \ding{55} & 67.4   & 44.3  &  \textbf{88.3}    &  65.1   &   93.4   &  83.7   &   62.6    &   23.7    &   42.0     &    63.6      \\
  
  Ensemble        & \ding{55} & 67.0   & 45.0  &  86.9    &  65.2   &   93.6   &  82.9   &   65.6    &   23.2    &   50.4     &    64.6      \\

  TPT            & \ding{55} &  \textbf{69.0}   & 47.8  &  87.8    &  68.0   &   \textbf{94.2}   &  \textbf{84.7}   &   65.5    &   24.8    &   42.4     &    65.1      \\

    APM (Ours)          & \ding{55} &  62.0   & \textbf{48.9}  &  81.6    &  \textbf{72.6}   &   89.6   &  84.2   & \textbf{65.7} &   \textbf{29.7}    &   \textbf{55.7}     &    \textbf{65.5}      \\
  
    \bottomrule
    \vspace{-1em}
     \label{tab:fine-grained}
  \end{tabular}
}
\end{table}

\textbf{APM is computationally efficient: } All experiments are run on a \textit{same} desktop-workstation containing 1x rtx a6000/96GB ram/Ubuntu 22.04/2TB ssd. Flops were counted with meta's fvcore library \cite{fvcore}. In Tab\ref{tab:compute_analysis}, we perform 20 TTT iterations. TTT on CLIP-VIT B/16 baseline used in \cite{tpt} consumes 462Gflops. Next, we do TTT leveraging APM. At $t=1$, feed-forwards involves clip-teacher and consumes $20.5$ flops. Remaining $19$ TTT iterations only involve overfitting on distilled image token at $t=1$, and consumes $10$ flops/TTT-iteration. The entire profile-dump yields $241.7$ flops, which is an almost $50\%$ reduction over $462$ flops. The total memory occupied by APM i.e. $2.7 GB$ is more than CLIP-VIT/B $2.3GB$ since teacher is kept in memory during TTT. However, APM only occupies $600MB$ and reduces actual consumed-flops by $50\%$.

\begin{table}[t]
\vspace{-1em}
\centering
\caption{\textbf{APM's computational analysis}: TTT for 20 iterations on APM. Baseline is CLIP VIT-B/16 which is used as a teacher in \cite{tpt}. $M_{meas}(GB)$, $GFlops_{meas}(GB)$ are tmeasured stats. $M_{i}(GB)$, $GFlops_{i}(GB)$ are idealistic estimates. (t): $t^{th}$ ttt iteration, (s): student, (u): teacher, (s+u): portion of memory/flops consumed by student/teacher respectively. Note that APM is a $25M$ net.  }
\label{tab:computational_analysis}
\begin{adjustbox}{width=0.9\textwidth}
\begin{tabular}{l*{6}{c}}
\toprule
& $t$ & Params(M)$\downarrow$ & $M_{meas}(GB)$ $\downarrow$& $M_{i}(GB)$ $\downarrow$& $GFlops_{meas}$ $\downarrow$& $GFlops_{i}$ $\downarrow$ \\
\midrule

Swin\cite{liu2021swin}  & 1-20 & 87 & 1.5 & 1.4 & 353 & 308 \\
\hline
TPT\cite{tpt}  & 1-20  & 151.3 & 3.1 & 2.7 & 529 & 476 \\
\hline
CLIP VIT-B/16 & 1-20  & 149.2 & 2.3 & 1.8 & 462 & 410 \\
\hline
CLIP VIT-B/16(u)\cite{tpt} & 1 & 149.2 & 1.8 & 1.8 & 20.5 & 20.5 \\
APM(s) & 1 & 174.2(s+u) & 2.7 (s+u) & 1.8(u) + 0.6(s) & 20.5(u) & 20.5(u) \\
APM(s) & 2 & 174.2(s+u) & 2.7 (s+u) & 1.8(u) + 0.6(s) & 10(s) & 10(s) \\
Peak Use & 1-20 & 174.2(s+u) & 2.7 (s+u) & 1.8(u) + 0.6(s) & \textbf{241.7 (s+u)} & \textbf{210.5 (s+u)} \\ 
\bottomrule
\vspace{-1em}
\label{tab:compute_analysis}

\end{tabular}
\end{adjustbox}
\end{table}

\begin{figure}[ht!]
    \centering
    \vspace{-1em}
    \includegraphics[width=0.9\linewidth]{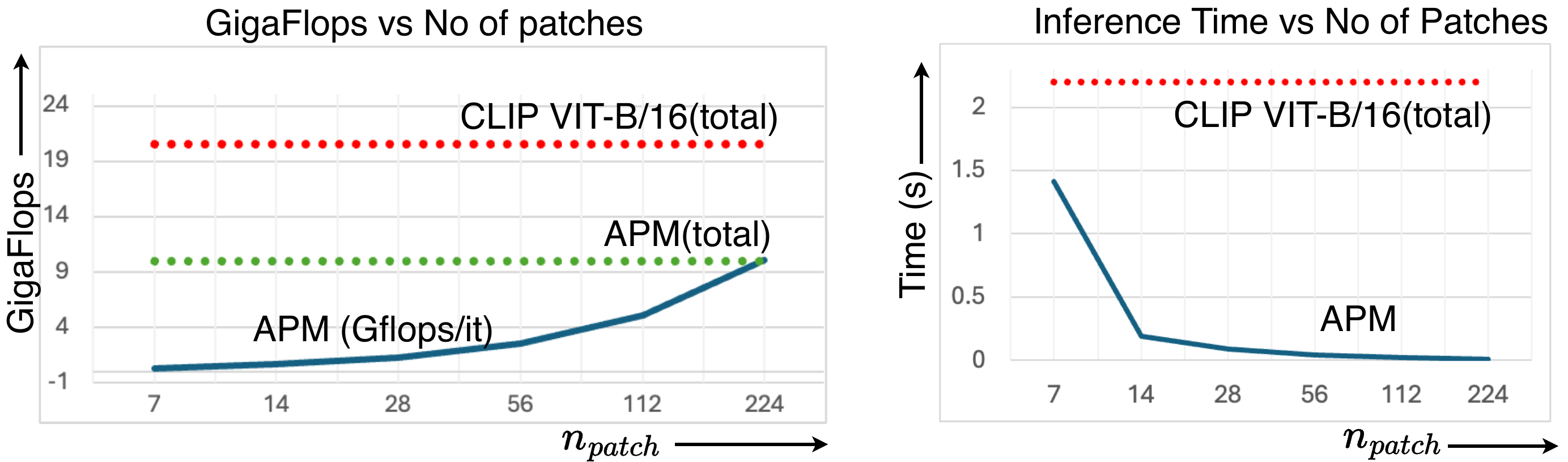}
        \caption{\textbf{APM's analysis with variable number of patches:} (left) Gflops of CLIP VIT-B/16 and APM as a function of number of processed patches. (right) Feed-forward time vs number of patches.      
        } 
    \label{fig:memory_gsraph}
     \vspace{-1.5em}
\end{figure}

\textbf{APM does patch-based processing:} In Fig \ref{fig:memory_gsraph}, we estimate GFlops/time to process a $224 \times 224$ image using CLIP VIT-B/16 \& APM. CLIP VIT-B/16 consumes 20.5 Gflops. On the other hand, APM takes only 10 Gflops in total, in part due to lesser parameters. APM's effectiveness comes from processing as low as 7 patches at the same time: it occupies lesser memory but takes more time (i.e. $1.5$ seconds). $1.5sec$ is still lower than VIT-B/16's $2.2 sec$. The extreme lies when all patches are processed: Inference time in APM goes down to $0.002$ seconds compared to VIT-B/16's $2.2$ seconds, thereby indicating its effectiveness. Note that VIT-B/16 can't do this patch-based processing. 

APM's computational effectiveness stems from this \textit{unique} ability to overfit on a \textit{single} test sample's embedding which was distilled only at $t=1$. This merits a deeper investigation.

\begin{figure}[ht!]
    \centering
    \vspace{-1em}
    \includegraphics[width=\linewidth]{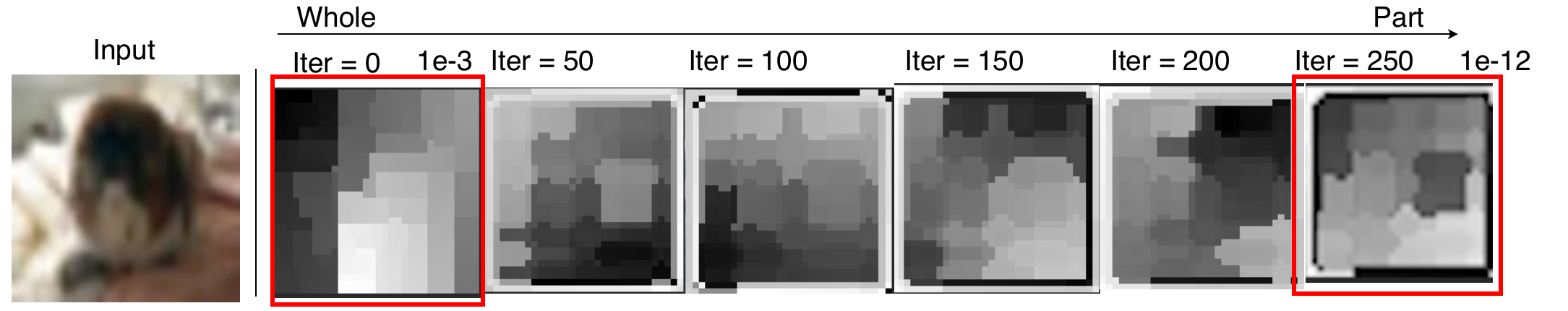}
        \caption{\textbf{Overfitting on a \textit{single} distilled token representation leads to islands of agreement\cite{hinton2023represent}:} APM is overfit on a test-sample's representation distilled from a teacher. We plot t-sne clustering of output features over 250ttt iterations. $L_2$ loss between predicted features  and distilled sample falls from 1e-3 to 1e-12. Moving left     to right shows that wholes break into smaller parts.         
        } 
    \label{fig:partwhole}
     \vspace{-1em}
\end{figure}

 \textbf{APM can learn on a single sample:} In Fig \ref{fig:partwhole}, we feed-forward a single image through our APM and perform $250$ TTT iterations\footnote{A larger number of TTT iterations are needed to perceive semantically-meaningful islands. Actual TTT process is far quicker, it takes around 20 iterations.}. We only \textit{overfit} on one distilled test-sample embedding and plot the 2D t-sne clustering of predicted APM. It can be seen that the elements of the scene have gradually started to cluster. This suggests that APM solves an \textit{inverse} problem: given a single test-sample embedding, what features in the image led to its formation? Over several TTT iterations, APM's features become representative enough to explain different parts of a scene\cite{hinton2023represent}. Therefore, the same-net \textit{could be made to move} up-down the part-whole hierarchy\cite{hinton2022forward}, although it requires TTT-iterations \textit{for now}.

Next, we explore if data augmentation \textit{improves} APM's performance. We perform  TTT iterations on  CIFAR-10's test set and find that data augmentation \underline{drops} APM's performance from $98.6$ to $76.7$. This quantitatively demonstrates that APM works best when it does \textit{one-sample} overfitting. It now aligns with the earlier qualitative experiment: over TTT-iterations, the network is learning to cluster the elements in the scene\ref{fig:partwhole}. Data augmentation \textit{distorts} the sample and makes it difficult for APM's predicted features to \textit{agree} on a stable, \textit{relaxed}-representation that explains the scene\cite{hinton2018matrix,hinton1976using}

Till now, APM's operation has involved random-initialization of weights for \textit{every} test-sample and performing test-time-training. There is another mode that it can be made to operate in.

\section{APM Training (Qualitative Analysis)}

APM can also scale up and do learning on a \textit{batch} of samples (for e.g., COCO images\cite{lin2014microsoft}) distilled from a teacher\cite{hinton2015distilling}. This requires introducing several new mechanisms. Note that this section is meant to \textit{qualitatively} demonstrate how scaling up APM can improve the net's interpretability, and help seed future research. Quantitative experiments beyond test-time-training remain \textit{out of the scope} of this paper. APM's training follows a standard setup in self-supervised-learning\cite{caron2021emerging}. We have provided the full algorithm for SSL-training/test-time-training in the Appendix\ref{sec:training_algorithm}. During inference, APM takes any 2D image $x_{k}$ and predicts its RGB reconstruction $RGB{k}$/higher dimensional features $f_{k}$. The net then begins to demonstrate several interesting properties, which we will discuss next.

\subsection{APM can do RGB reconstruction for any 2D input.}
\label{sec:rgb_symmetry}

\begin{figure}[ht!]
    \centering
    \vspace{-1.5em}
    \includegraphics[width=\linewidth]{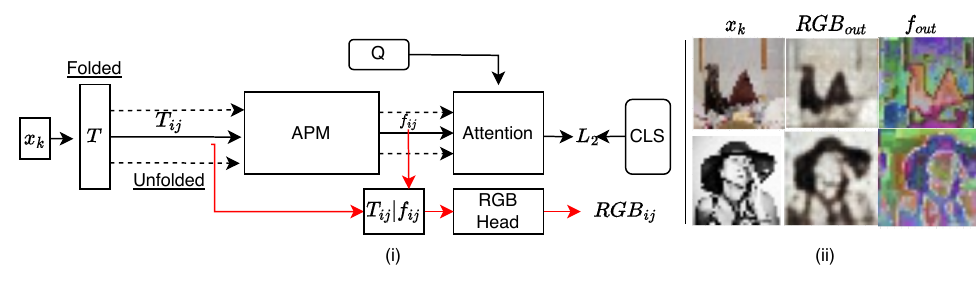}
    \vspace{-2em}
        \caption{ \textbf{RGB Decoding in APM}: Input trigger column  $T_{ij}$ is concatenated with predicted feature $f_{ij}$ and fed to downstream RGB head. This decodes RGB logit at location (i,j) for \textit{any} 2D input $x_k$. (ii) Input $x_k$ sampled from Coco-val set. $RGB_{out}$: reconstructed RGB, $f_{out}$: Predicted feature grid.}
    \label{fig:APM_routing}
    \vspace{-0.75em}
\end{figure}

Given a sample $x_{k}$, APM can reconstruct its RGB. In Fig\ref{fig:APM_routing}, we achieve this by estimating $f_{rgb,ij} = (T_{ij}| f_{ij})$. This skip-connection from trigger column $T_{ij}$ to output feature has a \textit{subtle} reason: consider a \textit{white} dog and \textit{brown} dog. The predicted object-level feature for both will be almost identical\cite{hinton2023represent,amir2021deep}. However, $T_{ij}$ is different for both since it contains lower patch-level features\cite{hinton2023represent}. Therefore, this helps us break symmetry. Without this skip-connection\cite{he2016deep}, the net fails to predict RGB. The network is trained to reconstruct RGB for a batch of images, $L_{rgb} = \sum_{i=1}^{N} \sum_{j=1}^{h*w} L_2(p'_{j}, p_{j})$, where $p_j$/$p_{j}^{'}$ are ground truth/predicted RGB-logits respectively. 

\subsection{APM is asynchronous yet encodes semantic-awareness in the net.}

\begin{figure}[ht!]
    \centering
    \includegraphics[width=\linewidth]{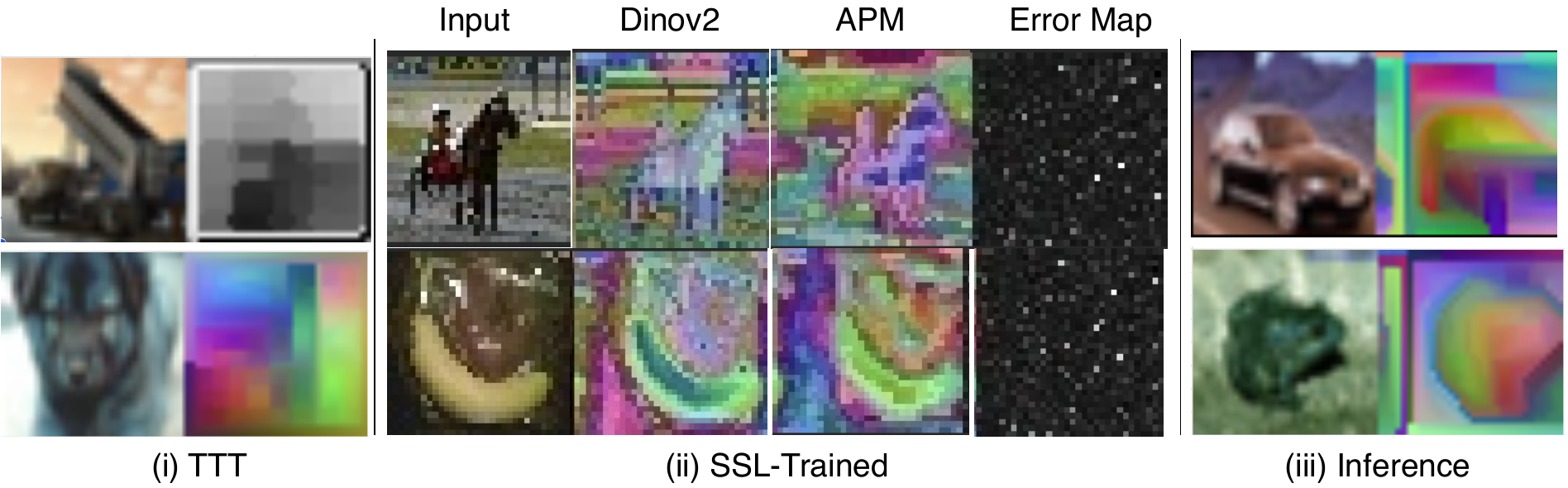}
    \caption{ 
   \textbf{APM feature Analysis:} (i) TTT iterations on an input image leads to semantically aware clustering. top: 2D t-sNE. bottom: 3D t-sNE. \cite{modi2023occlusions,hinton2023represent}. (ii) APM is trained via self-supervision using DINOv2-Teacher. (from left) Input, Dinov2 grid, APM grid. APM's grid  \textbf{closely approximates} Dinov2 grid evident from black regions in error map. Note that APM does asynchronous patch-based processing whereas Dinov2 does parallel perception. (iii) Cifar-10 samples feed-forwarded through SSL-trained APM yields features of significant semantic quality.\cite{hinton2023represent} }
    \label{fig:islands}
    \vspace{-2em}
\end{figure}

Given a sample $x_{k}$, APM can directly \textit{learn} to mimic the entire \textit{last layer} feature-grid which a teacher model would have generated. We enforce this by a $L_{grid} =  \sum_{i=1}^{N} \sum_{j=1}^{h*w} L_2(f'_{j}, f_{grid})$. In Fig \ref{fig:islands}(ii) we estimate the error map between the features predicted by our APM and Dinov2\cite{oquab2023dinov2}. The error map is mostly black which shows that it closely approximates Dinov2's grid\footnote{This particular net got a $L_{2}$ grid loss of $0.15$.}. Note that APM does patch-based \textit{asynchronous} processing whereas DINOv2 relies on parallel perception. Finally, Fig \ref{fig:islands}(iii) shows a simple feed-forward of a CIFAR-10 sample through APM. We can see semantically-aware features. Note that this is a \textit{single} feed-forward through APM\cite{hinton2023represent}. Predicting output feature grid allows the net to encode dark knowledge, i.e. the knowledge of both correct and \textit{incorrect} classes\cite{hinton2014dark}. This is better than just mimicking one hot vector of a target class since it manages to encode lower probabilities of wrong classes also.

\subsection{APM is a step towards validating GLOM's insight: input percept is a field\cite{hinton2014parallel}.}
\label{sec:continous_interpolation}

\begin{figure}[ht!]
    \centering
    \includegraphics[width=\linewidth]{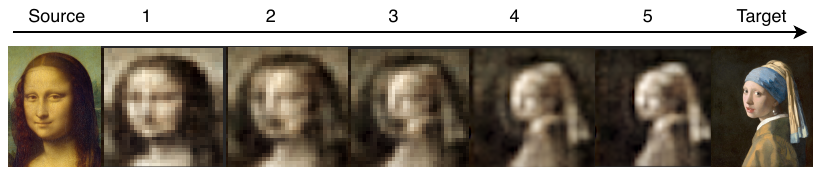}
    \caption{\textbf{APM is a step towards validating GLOM's insight \cite{hinton2023represent}}: input percept is a field. An interpolation between \textit{any} two images in the wild. This field arises in APM's MLP consisting of \textit{5 layers}. Trigger column $T$ acts as a key which retrieves an image from the APM's memory. T resides in a continuous embedding space, not discrete addressing space.}
    \label{fig:interpolation}
    \vspace{-1.em}
\end{figure}

In Fig \ref{fig:interpolation}, we show that APM can interpolate between any two images in the wild. We choose two images $I_1$ and $I_{2}$. These images are then funneled through the trigger column $T$ and yield two vectors $v_{1}$ and, $v_{2}$ respectively. Next, we generate n intermediate latents separated by an equal linear distance by $v_j = v_1 + \frac{v_2-v_1}{n}$. Each latent then brings into existence its own set of location-aware columns and decodes an image from the MLP. Such an interpolation has been previously observed in other models \cite{goodfellow2020generative}. APM now functions as a \textit{new form of addressing mechanism}: the trigger column T acts a key. \textit{Copying} T across locations yields image-specific queries\cite{hinton2023represent}. Values are synapses triggered in the MLP. RGB decoding happens in the output head. Hence, such \textit{continuous} keys and queries exist \textit{outside} the net\cite{hinton2023represent}. 

Classically, auto-regression has unrolled a shared-decoder over time\cite{werbos1990backpropagation}. In contrast, APM holds the whole sequence $I$ in $T$, and directly hops onto a space/time-step\cite{einstein1905electrodynamics} by querying the MLP with a location-column $T_{ij}$. Note that $T_{ij}$ is generated by unfolding $T$. Recurrence/feedback-loops are compensated for by a form of feature-expression\cite{hinton2023represent}. This is a step towards validating GLOM's insight, i.e. input-percept is a field\cite{hinton2023represent} and one can now interpolate in it (gestalt psychology). Furthermore, the trigger column $T$ resides in a continous embedding space, and not discrete hardware locations(classical AI)\cite{hinton2023represent}. Therefore, APM tries to integrate insights from both fields.

\section{Ablations on APM}
\label{sec:ablations}
\begin{table}[t!]
\footnotesize
\centering
\vspace{-1em}
\caption{\textbf{Ablations on APM}. All nets except CLIP VIT-L/14 use random weights b) $T_c$: trigger column contains convolutions. $T_{vit}$: Trigger column contains a routed VIT representation. C-10: CIFAR-10, C-100: CIFAR-100. Accuracy is reported. 
}
\label{tab:combined_small_tables}

\begin{subtable}[t]{.5\linewidth}
\caption{Ablation to evaluate abilities of existing nets to learn from a single sample\cite{he2016deep, rosenblatt1958perceptron}.}
\label{tab:overfit_ablation}
\setlength{\tabcolsep}{8pt}


\begin{tabularx}{\linewidth}{l|c|c|c}
              & Params & \multicolumn{1}{l|}{C-10} & \multicolumn{1}{l}{C-100} \\
\hline
Zeroshot      &        & \multicolumn{1}{l|}{}         & \multicolumn{1}{l}{}          \\

CLIP VIT-L/14 & 428M   & 95.37                        & 73.28                         \\
\hline
MLP           & 21M    & 9.0                          & 3.8                           \\
ResNet18      & 11.4M  & 85.69                        & 21.77                         \\
ResNet34      & 21.5M  & 78.24                        & 12.89                         \\
\hline
APM           & 25M    & \textbf{97.04}                        & \textbf{77.98}             
\end{tabularx}




\end{subtable}
\quad 
\begin{subtable}[t]{.45\linewidth}
\setlength{\tabcolsep}{6pt}
\caption{Ablations of our APM on C-10.$I_{xy}$ means that local patch was injected into the column $T_{c}$.}
\label{tab:ablation_triggercol}
\begin{tabularx}{\linewidth}{l|c|c|c|c|c}
        & $L_{grid}$ & $L_{cls}$ & $L_{rgb}$ & $I_{xy}$ & Acc \\
\hline
$T_{c}$ &  \xmark   & \cmark   & \xmark   & \xmark  & 94.2  \\
$T_{c}$ &  \xmark   &  \cmark  &  \cmark &\xmark   &  91.0 \\
$T_{c}$ &  \cmark   &  \cmark  & \xmark  & \xmark  & 96.1  \\
$T_{c}$ &  \cmark   &  \cmark  & \cmark  & \xmark  & 96.5  \\
$T_{c}$ & \cmark    &  \cmark  & \cmark  & \cmark  & \textbf{96.8}  \\
\hline
$T_{vit}$ & \cmark    &  \cmark  & \cmark  & \cmark  & \textbf{98.6}  \\
\end{tabularx}

\end{subtable}


\vspace{-2.5em}
\end{table}

The experiments on TTT had relied on a curious ability of APM: it could simply overfit on a test-sample's distilled representation at $t=1$. This merits further investigation:
        
    \textbf{Effect of one-sample test-time training:} In Table \ref{tab:overfit_ablation}, we investigate whether existing networks are capable of one-sample test-time-training. We employ a randomly-initialized network to overfit on a distilled test sample's token obtained in the first TTT iteration \cite{hinton2022forward}. Standard MLP achieves low accuracies of $9.0$ and $3.8$ on CIFAR-10 and CIFAR-100, respectively. Notably, an 11.4M parameter ResNet18 outperforms the larger ResNet34 with 21.5M parameters. A reason might be that too many parameters in ResNet34 gives it too many degrees of freedom\cite{hinton1987horizontal}: it finds it hard to overfit on one sample. Our APM demonstrates strong performance on both CIFAR-10 and CIFAR-100, surpassing the CLIP VIT-L/14 baseline. 
    
\textbf{Effect of various losses on APM:} We analyze each row in Table \ref{tab:ablation_triggercol}. Initially, only the CLS token from the teacher was distilled into our network, resulting in a yield of $94.2\%$. When we added RGB reconstruction loss for input, accuracy dropped to $91.0$, attributed to the difficulty of breaking RGB symmetry \cite{he2022masked}. Subsequently, mimicking the entire feature grid and CLS token from the teacher increased accuracy to $96.1$. Adding both $L_{rgb}$ and $L_{grid}$ further improved performance to $96.5$. $L_{grid}$ here refers to the last-layer of the teacher. Notably, while simple RGB reduction decreased performance ($91.0$), a combination of $L_{grid}$ and $L_{rgb}$ enhanced our network, as lower RGB and higher object features complement each other \cite{hinton2023represent}. Injecting local patch information $I_{xy}$ into the trigger column improved performance to $96.8$. Finally, routing output tokens from a single VIT layer into the trigger column $T_{vit}$ strengthened the column and improved performance.

\textbf{Effect of increasing number of convolution in T:} Increasing number of convolutional kernels from $1$ to $3$ improves from $96.08$ to $97.67$.

    These ablations reveal: 1) APM can do one-sample overfitting for a test sample, 2) It helps to have both local patch and a strong image representation in the trigger column T, and 3) Increasing the levels of part-whole supervision strengthens the net.


\section{Related Work}

\textbf{Prompting Approaches:} Prompting is a mechanism to adapt a foundational-model to a downstream task in a zero-shot manner\cite{madan2024foundation}. However, prompting typically requires well-designed hand-crafted prompts. Prompt-tuning methods consist of learnable prompts which enable a parameter-efficient approach to fine-tune a foundational-model. CoOp\cite{zhou2022learning} applied prompt-tuning to CLIP. However, CoOp\cite{zhou2022learning} is sensitive to OOD-data, which CoCoOp\cite{zhou2022conditional} in turn compensates for by conditioning the prompts on model inputs. Similarly, TPT\cite{tpt} optimizes a prompt to encourage consistent predictions across multiple augmented views of the same test sample, and uses confidence-selection to filter out noisy-predictions. Note that TPT\cite{tpt} performs prompt-tuning over ViT-B/16 but requires feed-forward through \textit{all the model-layers} for \textit{every} iteration. 

\textbf{Test-Time-Optimization:} Introduces the notion where a model adjusts its decision boundary dynamically during testing, for eg, improving robustness to distribution shifts. Test-time training (TTT) generally adds a self-supervised multi-task branch, and performs a SSL-task like rotation, or masked-reconstruction to adapt the network to the test-sample\cite{ttt_mae}. These approaches typically initialize the net with pre-trained weights, for eg, Imagenet before undergoing ttt-iterations on a downstream corrupted-dataset. An alternate line of work, for eg, TENT\cite{wang2020tent} proposes to minimize entropy of batch-wise test-samples. However, TENT\cite{wang2020tent} requires more than one test-sample to converge towards an optimal solution, whereas APM can also operate on one test sample. Another line of work adjusts internal batch-norm-stats of a network\cite{schneider2020improving}. However, this makes the network-architecture inflexible and requires more than one test sample for optimization. Several other papers following the original pioneering-TTT paper\cite{ttt_sun} have worked on different problem formulations, for eg, assuming access to
an entire dataset (e.g.  \cite{liu2021ttt++,r1,r2,r3,r5,r6,r7}) or a batch (e.g. TENT \cite{wang2020tent}) of test inputs. 

In contrast, APM evaluates on a each test-sample independently. Inductively, APM does not require any dataset specific pre-training, a pretext task or prompt tuning. Some approaches also use higher-parameterized transformer/diffusion models\cite{r1} making the optimization compute-intensive. However in APM, for TTT iteration $t>1$, feed-forward is done through only 25M parameter APM and not the 149.2M Clip VIT-B/16, resulting in computational efficient test-time-training(Fig \ref{fig:memory_gsraph}).  
 APM also inherits zero-shot behaviour from CLIP, which allows it to bypass training a separate dataset-specific linear-probe for downstream TTT. Finally, some works in areas like source-free domain-adaptation do perform TTT on smaller datasets like Cifar-10 etc\cite{wang2020tent}. APM additionally shows results on Imagenet splits (Tab\ref{tab:imagenet_splits},\ref{tab:imagenet_c}) and various cross-generalization datasets(Tab\ref{tab:fine-grained}).  

\textbf{Part-whole hierarchies:} The idea of encoding part-whole dynamic-parse-trees as distributed representations in neural nets can be traced back to \cite{hinton1990mapping}, with recent attempts leading to capsule networks\cite{hinton2018matrix,sabour2017dynamic}. However, the EM\cite{hinton2018matrix}/attention-based\cite{sabour2017dynamic} routing forces \textit{each} capsule to represent only one part\cite{modi2023occlusions,modi2017steganography, hinton2023represent}. This fundamental flaw prevents capsules from scaling-up and generalizing to multiple OOD objects\cite{modi2023occlusions}. Recently, GLOM\cite{hinton2023represent} proposed a \textit{theoretical} system of representing each input pixel as containing a column-vector. These vectors then undergo a routing procedure such that pixels corresponding to same part come to `agree' with each other.\cite{garau2022interpretable} demonstrated GLOM on Cifar splits. APM does not need any routing because it processes location-aware columns independently. APM now reveals an \textit{additional} perceptual-interpolation property not shown earlier(Fig\ref{fig:interpolation}). 

\textbf{Knowledge Distillation:} There has been a long history of training data-specific mixture/product of experts and having their ensemble vote towards a prediction, with the key idea to make the experts as different from as each other as possible, and only one expert deciding on the predictions of a particular input sample\cite{jacobs1991adaptive}. This is better than having all experts give equal opinion on a sample\cite{hinton2002training}. Other methods attempted to `gather' their collective knowledge to a single model for edge-device deployment\cite{hinton2015distilling}. Recently, knowledge has been transferred from a larger teacher model to smaller ones\cite{carion2020end}, for eg, a foundational model\cite{team2023gemini,qi2024interactive}. The student often retains/becomes-better than its teacher\cite{ahmad2023ez}. In practice, the weights of teachers are fluctuated slowly (EMA) as compared to the  student\cite{tarvainen2017mean}. This mechanism then helps realize Kahnman's theory of slow-fast thinking\cite{kahneman2011thinking}.

Typically, distillation requires boltzmann-matching\cite{boltzmann2022lectures} predicted distribution between students and teachers, with the distribution's sharpness being governed by a temperature parameter\cite{hinton1984boltzmann}. In contrast, APM directly mimics the entire last layer feature grid of a teacher via $L_{2}$ norm\cite{turki2023suds}. Furthermore, APM possesses a novel-inductive bias that \textit{can recover} semantic-features from a single CLS-token distilled from a teacher[Fig \ref{fig:islands}]. This validates the intuition that CLS tokens encode useful geometric-information of a scene \textit{after} cross-attention of CLS token with patch tokens of an input image\cite{dosovitskiy2020image}.  

\textbf{Routing Mechanisms}: Routing mechanisms involve routing correct object-specific information to correct neurons. \cite{locatello2020object} proposed slots which perform binding by iterative rounds of self-attention. Löwe et Al\cite{lowe2022complex} proposed 2D complex autoencoder and rotating features\cite{lowe2024rotating} where presence of an object is encoded in phase of a neuron and follows the minimum description length principle\cite{hinton1993keeping}. In APM, binding is done via the location itself\cite{hinton2023represent}. GroupVIT \cite{xu2022groupvit} routes information to multiple group tokens and shows that semantic segmentation emerges with just image-text contrastive supervision.

\section{Conclusion}
 
We propose APM, inspired from the insights presented in GLOM\cite{hinton2023represent,hinton1987learning}. APM \textit{promises} to be an efficient architecture for test-time-training and asynchronous patch-processing\cite{turing1990chemical}. APM shows robustness to extreme distribution shifts\cite{hendrycks2019benchmarking}. APM demonstrates that MLP's can be made to semantically cluster a given image.  We hope that APM will help \textit{inspire} further research on \textit{simpler} weight-sharing, lower-memory, higher-bandwidth efficient-nets \cite{hinton2022forward}. 

    \textbf{Limitations:} In this work, we have mainly-focused on image-classification. Furthermore, APM requires multiple TTT iterations for now, although they might be reduced by exploring pre-training on a source-dataset\cite{hinton2022forward}. APM can work on just 1 sample with randomly-initialized weights. However, it still requires a single CLS token which has been distilled from a teacher pre-trained on a large-scale dataset. Our preliminary-experiments have revealed that APM can still do RGB-reconstruction \textit{without} a teacher\cite{he2022masked}, showing potential that APM can be self-sufficient and independent.



\section{Acknowledgements}

This research is supported by the Intelligence Advanced Research Projects Activity (IARPA) via Department of Interior/ Interior Business Center (DOI/IBC) contract number 140D0423C0074. The U.S. Government is authorized to reproduce and distribute reprints for Governmental purposes notwithstanding any copyright annotation thereon. Disclaimer: The views and conclusions contained herein are those of the authors and should not be interpreted as necessarily representing the official policies or endorsements, either expressed or implied, of IARPA, DOI/IBC, or the U.S. Government. We also thank many other people and MLCollective whose support made this work possible.

\bibliographystyle{plainnat}

\bibliography{main}

\newpage

\appendix

\vskip 5mm
\begin{center}
    \hrule height 4pt
    \vskip 3mm
    {\LARGE Appendix:  Asynchronous Perception Machine for Efficient
    Test-Time Training \\ }
    \vskip 3mm
    \hrule height 1pt
\end{center}

\section{Broader Impact}
There are two main ideas that led us to Asynchronous Perception Machine\cite{hinton1984boltzmann,hinton2023represent}. The first idea is that instead of thinking of features as a cuboidal feature grid\cite{hinton2014features}, one can think of it as a column vector at each location\cite{hinton2023represent,modi2023occlusions}. This helped us learn one to one mapping between the input rgb patch and the vector at a particular location. It led to the net being able to process one patch at a time. 

The second idea is this notion of collapsing information into a single starting point. Our previous understanding was that this `collapse' leads to degeneracy\cite{hinton1984boltzmann}. In this paper, the information can be recovered from the starting point by copying it many times and breaking symmetry with positional encoding. By asking the right questions at the right place at the right time, a net can thus learn to express correct features\cite{turing1990chemical,hinton2023represent}. Although the information can `degenerate' to a single starting point, the net can still learn thanks to the strong positional-prior injected by such periodic-encodings\cite{vaswani2017attention}.

\section{Future Work} 
APM offers a \textit{fresh perspective} towards machine-perception: i.e. patches can now be lazily-processed one-by-one asynchronously\cite{hinton2023represent}. It shall be very-exciting to see APM's potential on dense-tasks, video-understanding\cite{modi2022video} and alternative testing-schemes i.e. few-shot scenarios or testing with a `batch-of-samples'. Another direction might involve making test-time-training faster/more-efficient by exploring alternate zeroth-order optimization techniques\cite{malladi2023fine,hinton2022forward}. Finally, APM contains a local-field which emerges as a consequence of folding-unfolding: this might have potential-applications to generative\cite{goodfellow2020generative,ho2020denoising,dhariwal2021diffusion}/dreaming\cite{mordvintsev2015inceptionism}/sleeping-machines\footnote{They are not always sleeping, they sometimes wake up too.\cite{hinton1995wake}.}\cite{crick1983function}.

\section{Pseudo-code for APM's operation.}
\label{sec:training_algorithm}
In Algorithm\ref{alg:apm_training}, we have inflated the entire pseudo-code to train APM beyond the applications of test-time-training. The idea is that given an input image $x_{k}$ APM can learn to predict its entire feature grid $f_{k}$ and its rgb logits $RGB_{k}$. First, the net inputs an image $x_{k}$. $x_{k}$ is then routed to a trigger column $T$. T then brings several columns $T_{ij}$ into existence. Each of these columns is fired into the MLP to yield location-aware features $f_{ij}$. The loss is collected for all locations and backpropagation then estimates the gradients required to update the parameters of the APM. In this entire process, there were no labels being used. The feature grids could have been any layer of a net like DINOv2. APM thus manages to learn a perception field within itself \cite{hinton2023represent}. It can be then frozen, and used as a computational equivalent of any feature-extractor for a downstream-perception task.

\begin{algorithm}[ht!]
\SetAlgoLined
\KwData{Input data $X$, Student $S_{APM}$, $\theta_{APM}$, Teacher $U$ (frozen), Learning Rate $\eta$}
\For{each epoch}{
    \For{each data point $x_k$ in $X$}{
        $f_{k} \leftarrow  U(x_{k})$ \;
        $T \stackrel{\text{CNN}}{\leftarrow}  x_k$ (create trigger col)\ref{sec:trigger}, $x_k \in \mathds{R}^{3 \times H \times W}$ \tcc*[r]{Route sample $x_{k}$ in column $T$\cite{hinton2023represent}}
        $T_{ij} \stackrel{\text{Unfold}}{\leftarrow}  T$ \ref{sec:query_existence}, $1 \leq i \leq H, 1 \leq j \leq W$ \tcc*[r]{ Create location-specific queries\cite{hinton2023represent}}
        \For{i in range(H)}{
            \For{j in range(W)}{
            $f^{'}_{ij} \leftarrow S_{APM}(T_{ij})$ \tcc*[r]{ Feed-Forward column $T_{ij}$ into APM and decode $f^{'}_{ij}$\cite{hinton2023represent}}
            $RGB^{'}_{ij} \leftarrow RGB_{Head}(S_{APM}(T_{ij}))$\ref{sec:rgb_symmetry}\tcc*[r]{ Decode lower-level $RGB^{'}_{ij}$\cite{hinton2023represent}}
            $L_{f} = L_2(f^{'}_{ij},f_{kij})$\;
            $L_{RGB} = L_2(RGB^{'}_{ij},x_{kij})$\;
            $L = L_{f} + L_{RGB}$\;
            Compute $\nabla_{\theta_{APM}} L$\;
            $\theta_{APM} \leftarrow \theta_{APM} - \eta \nabla_{\theta_{APM}} L$\;
            }
        }
        $T \stackrel{\text{fold}}{\leftarrow}  T_{ij}$\ref{sec:query_existence} \tcc*[r]{ Collapse all location-specific queries $T_{ij}$ into T\cite{hinton2023represent}}
        }
    }
 \caption{Training APM in a self-supervised manner using a teacher U.}
\label{alg:apm_training}
\end{algorithm}

\begin{algorithm}[ht!]
\SetAlgoLined
\KwData{Input data $X$, Student $S_{APM}$, $\theta_{APM}$, Teacher $U$ (frozen), Learning Rate $\eta$}

$ R_{gt} \leftarrow  U(\text{class}_{\text{name}})$ \tcc*[r]{Compute text representation of gt classes via U\cite{hinton2023represent}}
\For{each test sample $x_k$ in $X$}{
    $f_{k} \leftarrow  U(x_{k})$ \;
    $\theta_{APM} \leftarrow  \mathcal{N}(\mu,\sigma)$\tcc*[r]{Draw net's weights for appropriate initialization.\cite{glorot2010understanding,he2015delving}} \; 
    $f^{'} \leftarrow  \text{None}$ \;  
     
    \For{each iteration $t$}{
        $T \stackrel{\text{CNN}}{\leftarrow}  x_k$ (create trigger column)\ref{sec:trigger}, $x_k \in \mathds{R}^{3 \times H \times W}$ \tcc*[r]{Route sample $x_{k}$ in column $T$\cite{hinton2023represent}}
        $T_{ij} \stackrel{\text{Unfold}}{\leftarrow}  T$ \ref{sec:query_existence}, $1 \leq i \leq H, 1 \leq j \leq W$ \tcc*[r]{Create location-specific queries\cite{hinton2023represent}}
        $L \leftarrow  0$ \; 
        
        \For{i in range(H)}{
            \For{j in range(W)}{
                $f^{'}_{ij} \leftarrow S_{APM}(T_{ij})$ \tcc*[r]{Feed-Forward column $T_{ij}$ into APM \cite{hinton2023represent}}
                $f^{'} \leftarrow \text{statistical}_{\text{RunningAverage}}(f^{'}_{ij})$ \; 
                Estimate Loss $L += L_2(f^{'}_{ij},f_{kij})$\;
            }
        }
        
        Compute $\nabla_{\theta_{APM}} L$\;
        $\theta_{APM} \leftarrow \theta_{APM} - \eta \nabla_{\theta_{APM}} L$\;
        $T \stackrel{\text{fold}}{\leftarrow}  T_{ij}$\ref{sec:query_existence} \tcc*[r]{Collapse all location-specific queries $T_{ij}$ into $T$\cite{hinton2023represent}}
    }
    
    return $\text{pred} \leftarrow \text{contrastive}_{\text{classification}}(f^{'}, R_{gt})$ \;
}
\caption{Pseudo-Code for operation of APM during Test-Time-Training.}
\label{alg:apm_testing}
\end{algorithm}

Furthermore, we present the pseudo-code of APM for test-time-training in Algorithm\ref{alg:apm_testing}. First, the textual encoder of the teacher is used for estimating representations of each ground truth class. Then over multiple ttt-iterations, the predicted feature of APM, a.ka. $f$ is refined via statistical running-average\footnote{Averaging is similar to pooling in convolutional-nets and might lose important information over time. It might be helpful to explore temporally-weighed averages or alternate type of long-term memory-banks\cite{wu2022defense}.}. During each learning iteration, the trigger column $T$ is undergoing phases of folding-unfolding. Finally $f$ is being used to perform zero-shot-classification with prior-computed representations $R_{gt}$. 

A question may be posed on how to decide the optimal number of ttt iterations $t$ to achieve optimal performance. Indeed, one might build additional inductive-bias in a student (aka APM) to estimate when its own fantasies\cite{hinton1995wake}/predicted semantic-features are better than the teacher's and stop dynamically/recurse until kickoff. Notions on soft decision-making for higher-level cognition are subtly embedded in \cite{von1966theory,bengio2017consciousness}. 

\section{Reproducibility Statement} 
In order to ensure the reproducibility of our experiments, we have shared the model in supplementary during review process. The code, model weights shall be released post-review. APM can work with a single GPU like pascal in less than 2 GB of memory. It can also parallelize on a cluster containing 2 nodes of 8 A6000 amperes. We have provided  details of hyper-parameters used in test-time-training (Tab\ref{tab:hparams}), and precise details of each layer of APM(Tab \ref{tab:architecture}).

\section{Implementation Details}
\subsection{Architecture and Hyperparameters}
\textbf{Architecture:} We inflate APM's architecture in Table\ref{tab:architecture}. For the TTT experiments, APM consists of only a single convolutional layer, and 5 MLP layers. Additionally, APM consists of a feature projection head containing a single linear layer, and an optional RGB head. It maybe noted that the number of kernels in the convolutional filter is only 1. This creates a subtle issue: the RGB reconstruction in Fig\ref{fig:APM_routing} is black and white. This is because a single kernel loses RGB channel information. Put simply, assume a tuple of 3 numbers representing RGB values,  $<1,2,3>$ ,$<4,1,1>$. For a convolution operation with a single kernel assuming unit weights, the answer is $6$. If this $6$ gets injected in the net, it is equally certain that the input was $<1,2,3>$ or a  $<4,1,1>$ making reconstruction from the RGB head of APM difficult. We found that this symmetry-breaking issue could be noticeably fixed with $n_{kernels} \geq n_{channels}$, where $n_{channels}$ is the number of channels $c$ in the input. Historically, various other rotational/translational/mirror-symmetries have   played an important role in designing boltzmann machines\cite{sejnowski1986learning}. The architecture in Table \ref{tab:architecture} is then meant to showcase APM's potential to an extrema: how much can it do \textit{even with a single} convolutional filter?

\begin{table}[h]
    \centering
    \caption{\textbf{APM architecture for TTT}: with input dimensions $h,w,c$ and feature dimension $d_{p}$: dimensionality of positional encoding. $s$: stride of convolutional filter in encoder, $d_{c}$: dimension of the CLS token of teacher on which APM learns.}
    \label{tab:architecture}
    \resizebox{\linewidth}{!}{%
    \begin{tabular}{l@{\hskip 15pt}l@{\hskip 15pt}l@{\hskip 15pt}l@{\hskip 15pt}l@{\hskip 15pt}l}
    \toprule
       & Layer &  Feature Dimension  & $n_{kernels}$ & Stride & Padding  \\ 
      & &  \footnotesize{(H $\times$ W $\times$ C)} & & & \footnotesize{Input / Output}  \\
      \midrule
      & Input & $h$ $\times$ $w$ $\times$ $c$ \\
        \midrule
  Encoder     & Conv   & $h/s$ $\times$ $w/s$ $\times$ $d$ & 1& $s$ & 0 / 0  \\
                                              
    \midrule
    \multirow{8}{0.1\linewidth}{Decoder}    & Linear        &  $(d_{p} + d)$ * $4096$  & - & - & -  \\ 
    & Linear        &  $4096$ * $4096$  & - & - & -  \\ 
    & Linear        &  $4096$ * $4096$  & - & - & -  \\ 
    & Linear        &  $4096$ * $2048$  & - & - & -  \\ 
    & Linear        &  $2048$ * $1024$  & - & - & -  \\ 
        \midrule
  Feature Projection Head     & Linear   & $1024$ * $d_{c}$ & - & - & -  \\
        \midrule
  \multirow{8}{0.1\linewidth}{RGB-Head(optional) }    & Linear    &  $(d_{p} + d+ 1024)$ * $4096$  & - & - & -  \\ 
    & Linear        &  $1024*3$ * $256$  & - & - & -  \\ 
    & Linear        &  $256$ * $256$  & - & - & -  \\ 
    & Linear        &  $256$ * $3$  & - & - & -  \\ 
    \vspace{0.25em}\\
        \bottomrule
    \end{tabular}
    }
\end{table}

\textbf{Hyperparameters:} All hyper-parameters utilized for APM during test-time-training are detailed in \ref{tab:hparams}. We leveraged the seed 42 in most of our experiments, and also conducted experiments with multiple seeds. The weight matrices in APM were initilized with from a random distribution with $\mu=0$ and $\sigma= 0.01$. All the code has been written in Pytorch version 1.13.0. We also note that performing test-time-training with $16$ bit floating point allows us to effectively use recent GPU architectures for eg, Ampere: they contain a larger number of tensor cores \textit{in addition} to CUDA cores which results in significant speedups during the exprimentation process. Finally, we normalize an input image using standard Imagenet stats, and  \textit{dont resort to any other form of augmentation}, thereby making the pipeline far-simpler. 

\begin{table}[ht!]
    \centering
    \caption{\textbf{APM hyperparameters}  during test-time-training.}
    \label{tab:hparams}
    \begin{tabular}{ll}
    \toprule
    Number of Test samples & 50000 (Imagenet Splits), variable for other datasets.\\
    Testing iterations         & 20      \\
        Batch Size              & 1        \\
        Learning Rate         & 1e-4       \\
        Optimizer         & Adam   
         \\
    \midrule
        Feature Output size $d$         & $768/1024$       \\
        Positional  Encoding size &$768/1024$\\
    \midrule
        Image/Crop Size         & 448        \\
        Augmentations           & Normalization,  $\mu = (0.485, 0.456, 0.406)$, $\sigma= (0.229, 0.224,
0.225)$      \\
        
    \midrule
    Precision & fp16 (grad-scaled) \\
    Num of Workers & 8 \\
    Operating System & 1x rtx a6000 48GB/96GB ram/Ubuntu 22.04/2TB ssd/5TB HDD\\
    \bottomrule
    \end{tabular}
    \vspace{-2em}
\end{table}

\paragraph{ViT Encoder:} During our experiments in test-time-training, APM relies on higher-dimensional CLS token distilled from a teacher trained on a large-scale-dataset, often via contrastive image-text objectives. We showed quantitative results with CLIP, OpenCLIP and qualitative semantic-clusterings with DinoV2. 

CLIP is a zero-shot model from OpenAI which contains a vision encoder, and a textual encoder. The textual encoder tokenises input class names to features. Both image/text encoder project them to common dimensionality, and classification happens by measuring distances in contrastive space, thereby offering a higher degree of freedom, as opposed to training a class-sensitive linear-probe. CLIP VIT-L features an output CLS token of $768$ dimensions, while CLIP VIT-H outputs $1024$ dimensions both of which have been accommodated in Tab\ref{tab:architecture}. DinoV2\cite{oquab2023dinov2} is a foundational-model trained via SSL-objectives and predicts significant semantically-aware representations, which are widely used in various downstream computer-vision tasks. 

Inductively, both CLIP/Dinov2 rely on VIT, which operates on the principle of parallel attention\cite{vaswani2017attention}: image-tokens a.k.a patches can flow along parallel paths among different layers stacked over each other without any loss of spatial-resolution which was also a key-shortcoming of convolutional-nets. Even though the transformers have no bottleneck issue, the attention-operation in each layer still occupies a significant amount of memory\cite{modi2023occlusions}.

\subsection{Datasets:}

To evaluate model's robustness to distribution shifts, it is necessary to test them on datasets which contain corruptions on a variety of scenarios for eg, \textit{fog, rain, snow, etc.}. One standard practice has been to take the test set of larger datasets like ImageNet, and create synthetic corruptions to establish appropriate-benchmarks on which the performance of these models can be compared. Alternatively, certain test splits have  been manually-curated \textit{from-scratch} over the internet, for eg, sketches, artistic-drawings etc. Below, we detail some of the splits which were used in this paper for APM's robustness experiments.

\textbf{Cifar-10C}: is a test-split consisting of $10000$ test-samples of Cifar-10, corrupted with $15$ noises, across $5$ levels of noise-severities. In this paper, we have shown results on the highest level, a.k.a level 5 owing to the resource-constraints. 

\textbf{Imagenet-C:} ImageNet-C is a dataset split for recognizing objects under distribution shifts, with 1000 classes like original Imagenet. This split contains 15 types of corruptions, with each type containing 5 level of noise severities, aka, the percentage of the image region which is being typically impacted by the corruption.

\textbf{ImageNet-V2}: is an independent test set containing images sampled from naturally occuring scenarios, with 10000 images of 1000 ImageNet classes. ImageNet-V2 typically consists of 3 splits, with varying levels of difficulties.

\textbf{ImageNet-A}: refers to a curated test-set containing "natural occurring adversarial samples", which were misclassified by the Resnet-50. This particular split contains 7500 images of 200 ImageNet categories.

\textbf{ImageNet-R}:  refers to a novel test-set of several Imagenet categories, which contain artistic renditions. There are 30000 images in this split spanning across 200 ImageNet categories.

\textbf{ImageNet-Sketch}: is a challenging test-split which consists of only black-and-white sketches of 1000 ImageNet categories. This split originally consists of 50,000 images in total.

Typically, methods like CLIP evaluate on these ImageNet using a prompt-ensemble of $80$ handcrafted templates. APM results were also shown using this ensemble for fair comparisons.  

\section{Additional Ablations}
Here we perform some additional ablations to understand how varying one of the parameters of APM, while holding others hyper-parameters constant impacts the performance during test-time-training. 

\textbf{Effect of varying number of ttt iterations:} We perform varying number of ttt iterations, evaluate the performance of APM on three seeds, and report the mean and standard deviations. We note that at $t=25$, APM obtains $49.5\%$ accuracy with a minimum observed standard-deviation of $0.2$. 


\begin{table}[ht!]
\caption{\textbf{Ablation with variable ttt iterations on DTD dataset}: We pick the best performing 53M param net in tab 7 of the main paper. We show mean/std over 3 runs with seeds 0/7/42. The net settles on the best result of 49.5 and std reduces to 0.2 at $n_{iter}=25$.}
\label{tab:ablation_params}
\vspace{5pt}
\centering
\resizebox{0.6\linewidth}{!}{%
\begin{tabular}{lcccccc}
\toprule
$n_{iter}$ & 10 & 15 & 20 & 25 & 30 & 35 \\
\midrule
APM (53M net) & 44.2/0.6 & 47.7/0.7 & 49.1/0.5 & \textbf{49.5/0.2} & 48.3/0.6 & 46.3/0.1 \\
\bottomrule
\end{tabular}}
\end{table}

\textbf{Effect of varying number of parameters:} In Tab\ref{tab:ablation_params}, we perform an additional ablation: the number of parameters inside APM's linear layers are gradually changed from 7M to 120M. We perform TTT iterations on the DTD dataset. As can be observed, the top-1 classification-accuracy of APM increases from $47.0$ to $49.1$, thereby indicating that a $53M$ net was optimal for this particular instance of the problem. Beyond that, we observe a gradual drop in the performance, thereby indicating that the net has started to overfit. In an ideal scenario, we would want that lower number of parameters should yield higher performance. However, this would then require some more fundamental changes which allows the net to achieve higher-bandwidths, which remains a matter for the future work\cite{hinton2022forward}. 

\begin{table}[ht]
\caption{\textbf{Ablation on APM parameter count on DTD dataset}: Increasing the number of parameters to 53M improves APM's performance to 49.1 beyond which it starts to drop. Top 1 classification accuracy is being reported.}
\label{tab:ablation_params}
\vspace{5pt}
\centering
\resizebox{0.6\linewidth}{!}{%
\begin{tabular}{lcccccccc}
\toprule
& 7M & 25M & 36M & 53M & 70M & 87M & 104M & 120M \\
\midrule
APM & 47.0 & 47.5 & 48.1 & \textbf{49.1} & 48.4 & 48.1 & 47.8 & 47.0 \\
\bottomrule
\end{tabular}}
\end{table}

\textbf{Effect of removing the teacher from APM:} APM can perform test-time-training on a single test-sample by relying on a CLS-token distilled from a teacher trained on-scale. This might lend itself to the assumption that APM \textit{really requires} a teacher in order to learn semantically-aware representations. In this ablation, we remove the teacher entirely and have APM perform RGB-reconstruction on COCO-train set. The $L_2$ RGB-reconstruction loss on COCO-val loss then falls to 0.0027. Training took far longer than if last-layer feature-vectors were also distilled from a teacher into APM. 

Higher-dimensional vector-spaces carry more bits\cite{shannon1948mathematical}, but incur significant randomness\cite{boltzmann2022lectures}. Island/vector-distillation speeds-up the learning-process which otherwise might only be informed via RGB-reconstruction and take a lot of time\cite{he2022masked,hinton2023represent}. This helps guide APM's ship to correct points in the subspace. Progressing from VIT b->h in Tab\ref{tab:imagenet_splits}, shows APM becomes more competitive thereby validating this intuition.    

\section{Additional Results}

\textbf{Results on Cifar-10C:} In Tab\ref{tab:cifar_10_c},  we show additional results on Cifar 10-C dataset widely used in test-time-training \cite{ttt_sun}. Cifar-10C consists of $15$ types of noise corruptions for all the 10,000 samples present in the Cifar-10 dataset. As evident from the table, CLIP is \textit{not} naturally robust on Cifar-10 and achieves an error rate as high as $24.5\%$. This is worse than existing ttt-methods like TTT-Online and UDA-SS. Therefore, CLIP VIT-L is \textit{not} naturally-robust on Cifar-10c.

Using it as a teacher, we get the \textit{lowest} average error rate of $14.8\%$, thereby even improving upon the performance of CLIP VIT-L/14. Note that our APM is reinitialized with \textit{random} weights after every-test sample in contrast to methods like TTT-Online which $retain$ the weights after every ttt-iteration. Inspite of that, we get a lower error rate i.e. $14.8\%$ than TTT-Online $19.1\%$. Another benefit which APM gains is that it can directly use the textual-encoder of the Clip VIT-L/14 teacher to classify on cifar-10 test set: this allows us to \textit{bypass} the requirement of training a separate dataset-specific linear probe on top of our net. 

\begin{table}[ht]
  \caption{\textbf{Cifar 10-C} results at \textit{highest} severity level of 5. We report \textbf{Error Rate}. Lower numbers are better. t- model acts as teacher for our APM. TTT was done on test set with randomly initialized weights. APM weights were reinitialized after each TTT iteration to prevent information leakage. Lower is better.}
  \label{tab:cifar_10_c}
  \centering    
  \resizebox{\textwidth}{!}{%
  
  \begin{tabular}{l|c|c|c|c|c|c|c|c|c|c|c|c|c|c|c|c|c}
  \toprule
Method     & orig & gauss & shot & impul & defoc & glass & motn & zoom & snow & frost & fog  & brit & contr & elas & pixel & jpeg & Avg \\
\hline 
TTT-Online & 8.2  & 25.8  & \textbf{22.6} & 30.6  & 14.6  & 34.4  & 18.3 & 17.1 & 20.0 & 18.0  & 16.9 & 11.2 & 15.6  & \textbf{21.6} & 18.1  & 21.2 & 19.1 \\
UDA-SS     & 9.0  & 28.2  & 26.5 & 20.8  & 15.6  & 43.7  & 24.5 & 23.8 & 25.0 & 24.9  & 17.2 & 12.7 & 11.6  & 22.1 & 20.3  & 22.6 & 21.4 \\
\hline
Zeroshot     \\
Clip VIT-L/14     & 4.63  & 35.4  & 32.3 & 21.9  & 19.3  & 49.7  & 19.3 & 17.3  & 17.0 & 15.1  & 21.6 & 8.4 & 15.9  & 34.6 &  25.0  & 27.4  & 24.5 \\
\hline 
Clip VIT-L/14 (t) \\
APM (Ours)    &  \textbf{3.5} &\textbf{ 21.9}  & 30.1 & \textbf{13.7}  & \textbf{15.2} & \textbf{34.1} & \textbf{11.9} & \textbf{11.1} & \textbf{15.0} &  \textbf{9.0} & \textbf{13.5} & \textbf{5.8} & \textbf{9.5}  & 23.0& \textbf{15.8}  & \textbf{17.0} & \textbf{14.8} \\
\hline
\end{tabular}
}
\end{table}

\textbf{Results on ImageNet-C:} In Tables\ref{tab:imagenet_c_level_1},\ref{tab:imagenet_c_level_2},\ref{tab:imagenet_c_level_3},\ref{tab:imagenet_c_level_4}, we perform test-time-training on APM for different imagenet splits across increasing levels of noise-severities. We observe that APM continues to obtain competitive performance over it's teacher. 


\begin{table}[ht]
\caption{\textbf{APM's performance on ImageNet-C, level 1}. The first two rows are same as the supplementary materials of \cite{ttt_mae}. A \ding{51} in P means that method leveraged \textbf{pre-trained weights} on clean variant of train set aka, Image-net and downstream-ttt on corrupted version. OpenCLIP VIT-L/14 is in general more robust. APM can surpass OpenCLIP VIT-L/14.}
\label{tab:imagenet_c_level_1}
\centering    
\resizebox{\linewidth}{!}{%
\begin{tabular}{l*{18}{c}}
\toprule
& P & brigh & cont & defoc & elast & fog & frost & gauss & glass & impul & jpeg & motn & pixel & shot & snow & zoom & Average \\
\midrule
\rowcolor[HTML]{EFEFEF}
Baseline & \ding{51} & 78.5 & 74.5 & 68.1 & 73.9 & 70.5 & 70.6 & 74.8 & 68.6 & 72.3 & 73.0 & 75.2 & 75.9 & 73.6 & 69.3 & 63.7 & 71.4 \\
\rowcolor[HTML]{EFEFEF}
TTT-MAE & \ding{51} & 78.9 & 74.7 & 72.5 & 74.7 & 72.9 & \textbf{72.2} & 76.8 & 72.2 & 75.5 & 74.5 & 75.8 & 77.0 & 75.9 & 71.9 & \textbf{69.3} & 73.1 \\
\midrule 
OpenCLIP VIT-L/14 & \ding{55} & 77.3 & 75.4 & 73.5 & 73.1 & 73.5 & 71.4 & 71.9 & 70.2 & 69.9 & 75.1 & 73.7 & 74.2 & 71.9 & 71.2 & 65.2 & 71.1 \\
APM (Ours) & \ding{55} & \textbf{81.6} & \textbf{80.3} & \textbf{78.6} & \textbf{78.0} & \textbf{78.6} & 76.6 & \textbf{77.2} & \textbf{75.7} & \textbf{75.1} & \textbf{79.6} & \textbf{78.7} & \textbf{79.1} & \textbf{76.9} & 76.4 & 70.7 & \textbf{76.0} \\
\bottomrule

\end{tabular}
}
\end{table}


\begin{table}[ht]
\caption{\textbf{APM's performance on ImageNet-C, level 2}. The first two rows are same as the supplementary materials of \cite{ttt_mae}. A \ding{51} in P means that method leveraged \textbf{pre-trained weights} on clean variant of train set aka, Image-net and downstream-ttt on corrupted version. OpenCLIP VIT-L/14 is in general more robust. APM can surpass OpenCLIP VIT-L/14.}
\label{tab:imagenet_c_level_2}
\centering    
\resizebox{\linewidth}{!}{%
\begin{tabular}{l*{18}{c}}
\toprule
& P & brigh & cont & defoc & elast & fog & frost & gauss & glass & impul & jpeg & motn & pixel & shot & snow & zoom & Average \\
\midrule
\rowcolor[HTML]{EFEFEF}
Baseline & \ding{51} & 77.4 & 71.2 & 62.3 & 51.0 & 66.3 & 58.4 & 68.6 & 59.2 & 64.9 & 70.4 & 70.6 & 74.7 & 66.2 & 54.2 & 55.2 & 64.1 \\
\rowcolor[HTML]{EFEFEF}
TTT-MAE & \ding{51} & 77.8 & 71.5 & \textbf{69.4} & 49.7 & 69.8 & 62.7 & 72.5 & 66.4 & 70.0 & 72.7 & 72.3 & 76.2 & 70.6 & \textbf{58.7} & 63.6 & 68.3 \\
\midrule 
OpenCLIP VIT-L/14 & \ding{55} & 76.6 & \textbf{74.4} & 71.4 & 53.8 & 72.0 & 62.6 & 67.6 & 64.0 & 64.6 & 73.8 & 69.0 & 72.8 & 66.4 & 61.8 & 58.3 & 66.1 \\
APM (Ours) & \ding{55} & \textbf{81.1} & \textbf{79.4} & 76.6 & \textbf{59.4} & \textbf{77.3} & \textbf{68.2} & \textbf{73.1} & \textbf{70.0} & \textbf{70.3} & \textbf{78.6} & \textbf{74.5} & \textbf{77.8} & \textbf{72.0} & 67.8 & \textbf{64.3} & \textbf{72.4} \\
\bottomrule
\end{tabular}
}
\end{table}


\begin{table}[h!]
\caption{\textbf{APM's performance on ImageNet-C, level 3}. The first two rows are same as the supplementary materials of \cite{ttt_mae}. A \ding{51} in P means that method leveraged \textbf{pre-trained weights} on clean variant of train set aka, Image-net and downstream-ttt on corrupted version. OpenCLIP VIT-L/14 is in general more robust. APM can surpass OpenCLIP VIT-L/14.}
\label{tab:imagenet_c_level_3}
\centering    
\resizebox{\linewidth}{!}{%
\begin{tabular}{l*{18}{c}}
\toprule
& P & brigh & cont & defoc & elast & fog & frost & gauss & glass & impul & jpeg & motn & pixel & shot & snow & zoom & Average \\
\midrule
\rowcolor[HTML]{EFEFEF}
Baseline & \ding{51} & \textbf{75.8} & 62.7 & 49.5 & 67.1 & 59.8 & 47.6 & 57.1 & 35.0 & 57.4 & 68.6 & 60.2 & 70.1 & 54.3 & 54.7 & 48.0 & 57.6 \\
\rowcolor[HTML]{EFEFEF}
TTT-MAE & \ding{51} & \textbf{75.8} & \textbf{64.4} & 59.4 & 71.2 & 64.0 & 54.0 & 63.6 & 50.7 & 64.2 & 71.3 & 64.2 & 73.1 & 61.8 & \textbf{58.0} & 57.4 & 64.4 \\
\midrule 
OpenCLIP VIT-L/14 & \ding{55} & 75.8 & 71.8 & \textbf{65.5} & 67.7 & 69.0 & 54.7 & 58.9 & 42.4 & 59.5 & 72.8 & 59.9 & 69.7 & 58.2 & 63.5 & 51.8 & 62.5 \\
APM (Ours) & \ding{55} & \textbf{80.5} & \textbf{77.2} & 71.3 & \textbf{73.3} & \textbf{74.8} & \textbf{60.6} & \textbf{64.7} & \textbf{48.5} & \textbf{65.4} & \textbf{77.8} & \textbf{61.6} & \textbf{75.2} & \textbf{64.1} & \textbf{69.3} & \textbf{58.0} & \textbf{68.5} \\
\bottomrule
\end{tabular}
}
\end{table}


\begin{table}[h!]
\caption{\textbf{APM's performance on ImageNet-C, level 4}. The first two rows are same as the supplementary materials of \cite{ttt_mae}. A \ding{51} in P means that method leveraged \textbf{pre-trained weights} on clean variant of train set aka, Image-net and downstream-ttt on corrupted version. OpenCLIP VIT-L/14 is in general more robust. APM can surpass OpenCLIP VIT-L/14.}
\label{tab:imagenet_c_level_4}
\centering    
\resizebox{\linewidth}{!}{%
\begin{tabular}{l*{18}{c}}
\toprule
& P & brigh & cont & defoc & elast & fog & frost & gauss & glass & impul & jpeg & motn & pixel & shot & snow & zoom & Average \\
\midrule
\rowcolor[HTML]{EFEFEF}
Baseline & \ding{51} & \textbf{73.1} & 33.1 & 35.8 & 56.9 & 54.2 & 45.2 & 39.6 & 26.0 & 38.2 & 62.0 & 43.2 & 60.3 & 32.2 & 44.2 & 40.7 & 47.4 \\
\rowcolor[HTML]{EFEFEF}
TTT-MAE & \ding{51} & 72.7 & 39.6 & 45.7 & 64.9 & 58.3 & 52.6 & 48.5 & 42.8 & 47.6 & 67.0 & 50.5 & 66.6 & 42.4 & 45.7 & \textbf{51.5} & 53.2 \\
\midrule 
OpenCLIP VIT-L/14 & \ding{55} & 74.2 & \textbf{64.2} & 58.7 & 57.8 & \textbf{66.3} & 52.8 & 45.3 & 34.6 & 45.2 & 68.9 & 46.6 & 63.9 & 41.1 & 56.2 & 45.6 & 54.8 \\
APM (Ours) & \ding{55} & \textbf{79.2} & 70.4 & \textbf{64.9} & \textbf{63.7} & 72.3 & \textbf{58.6} & \textbf{51.2} & \textbf{40.4} & \textbf{51.3} & \textbf{74.1} & \textbf{53.0} & \textbf{70.0} & \textbf{46.7} & \textbf{62.5} & 51.8 & \textbf{59.6} \\
\bottomrule
\end{tabular}
}
\end{table}

\begin{figure}[ht!]
    \centering
    \includegraphics[width=0.8\linewidth]{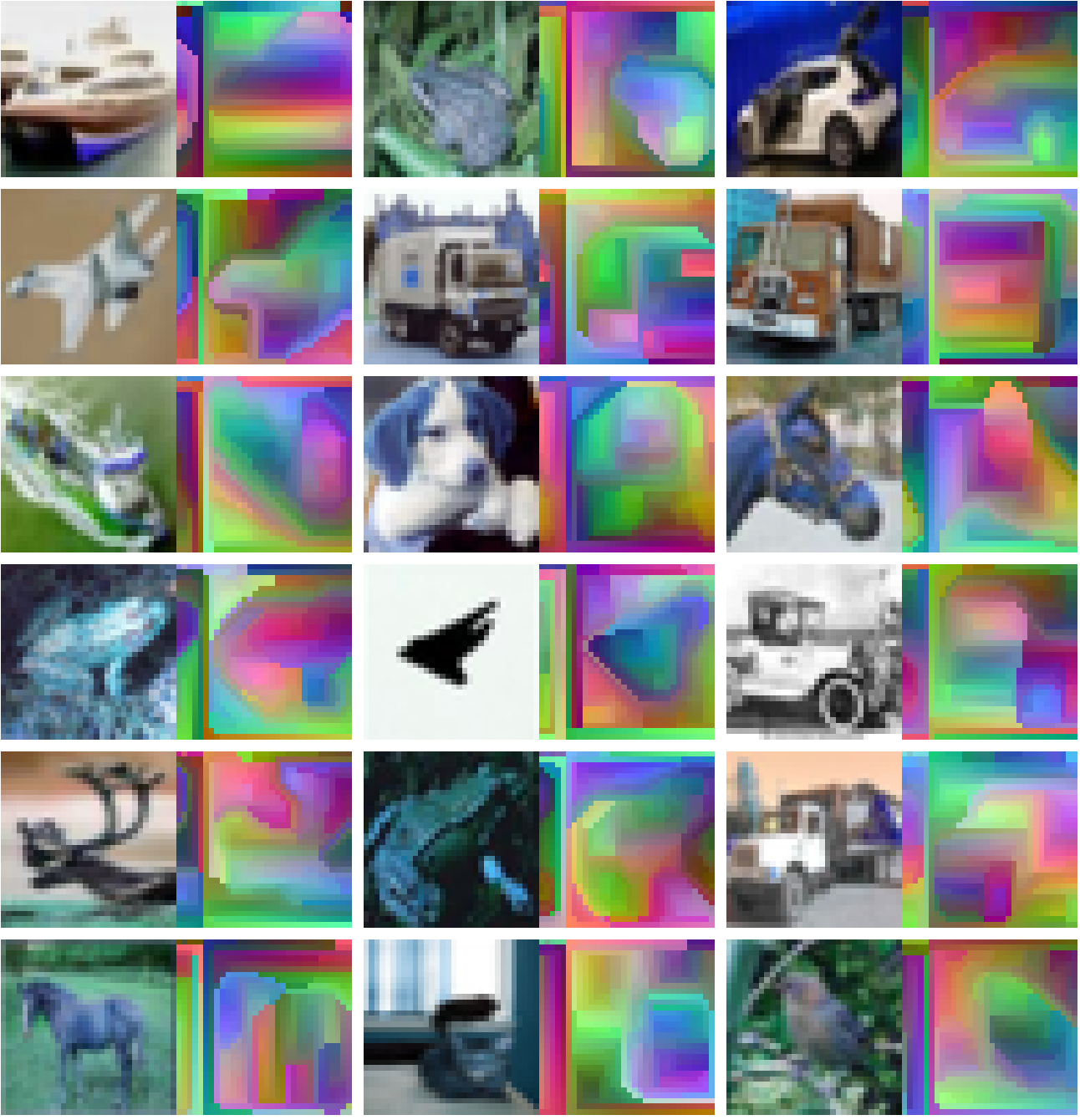}
    \caption{\textbf{Cifar 10 islands:} Individual part-wholes are clearly observed in APM features. These features are used for downstream classification.We leverage the visualization mechanism by \cite{modi2023occlusions}. These islands are \textit{not} been hand-picked\cite{hinton2023represent} }
    \label{fig:cifar_1}
    \vspace{-1em}
\end{figure}

\begin{figure}[ht!]
    \centering
    \includegraphics[width=0.8\linewidth]{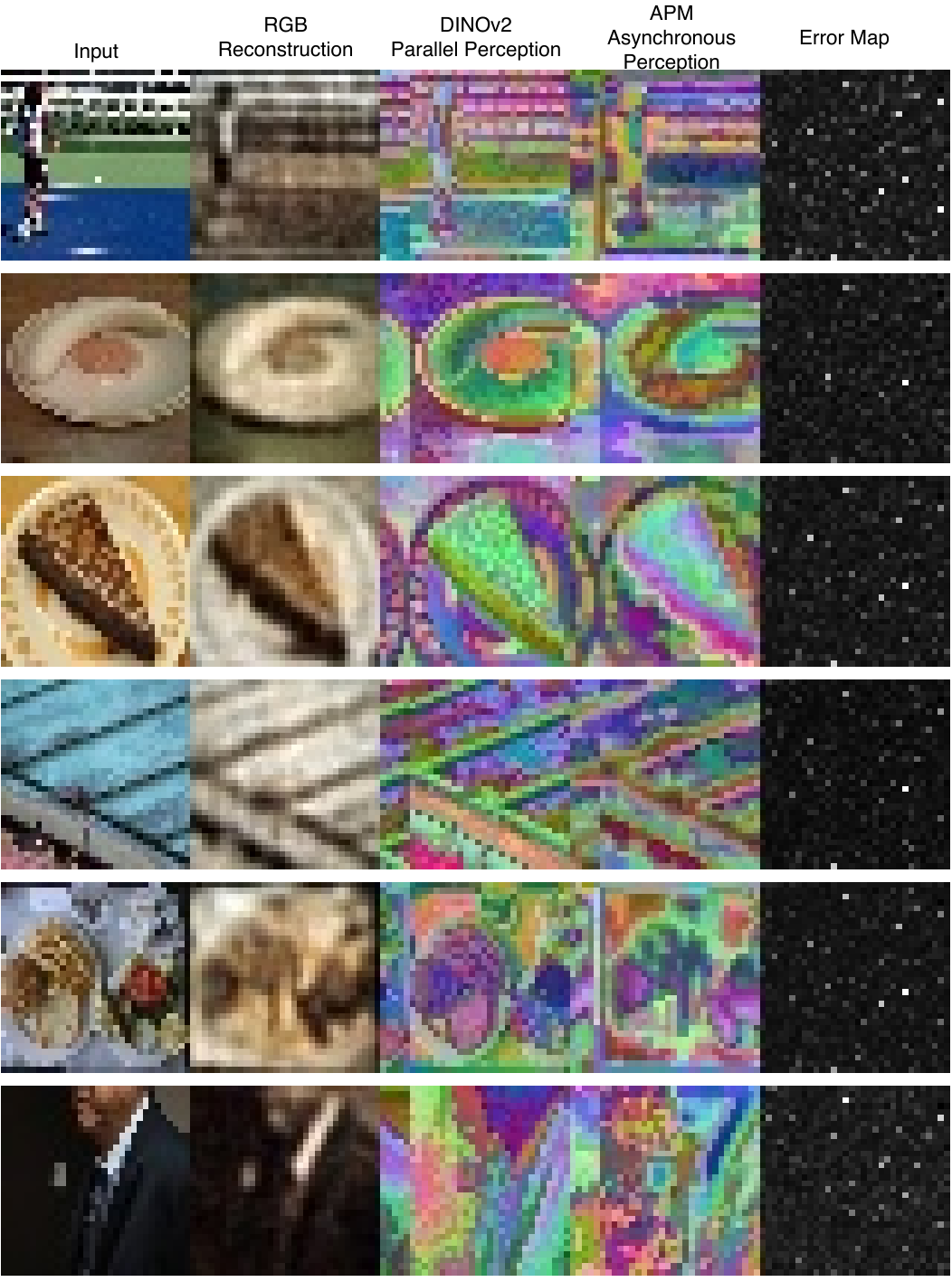}  
    \caption{\textbf{Qualitative Results on COCO Val set:} Our APM is trained on COCO-train set, and islands on COCO-val set are visualized\cite{hinton2023represent}. We leverage the visualization mechanism by \cite{modi2023occlusions}. Note that these islands are a consequence of \textit{single} feed-forward through the net and not an iterative routing mechanism as in \cite{garau2022interpretable}. These islands are \textit{not} been hand-picked.\cite{hinton2023represent} }
    \label{fig:coco_3}
    \vspace{-2em}
\end{figure}

\begin{figure}[ht!] 
    \centering
    \includegraphics[width=\linewidth]{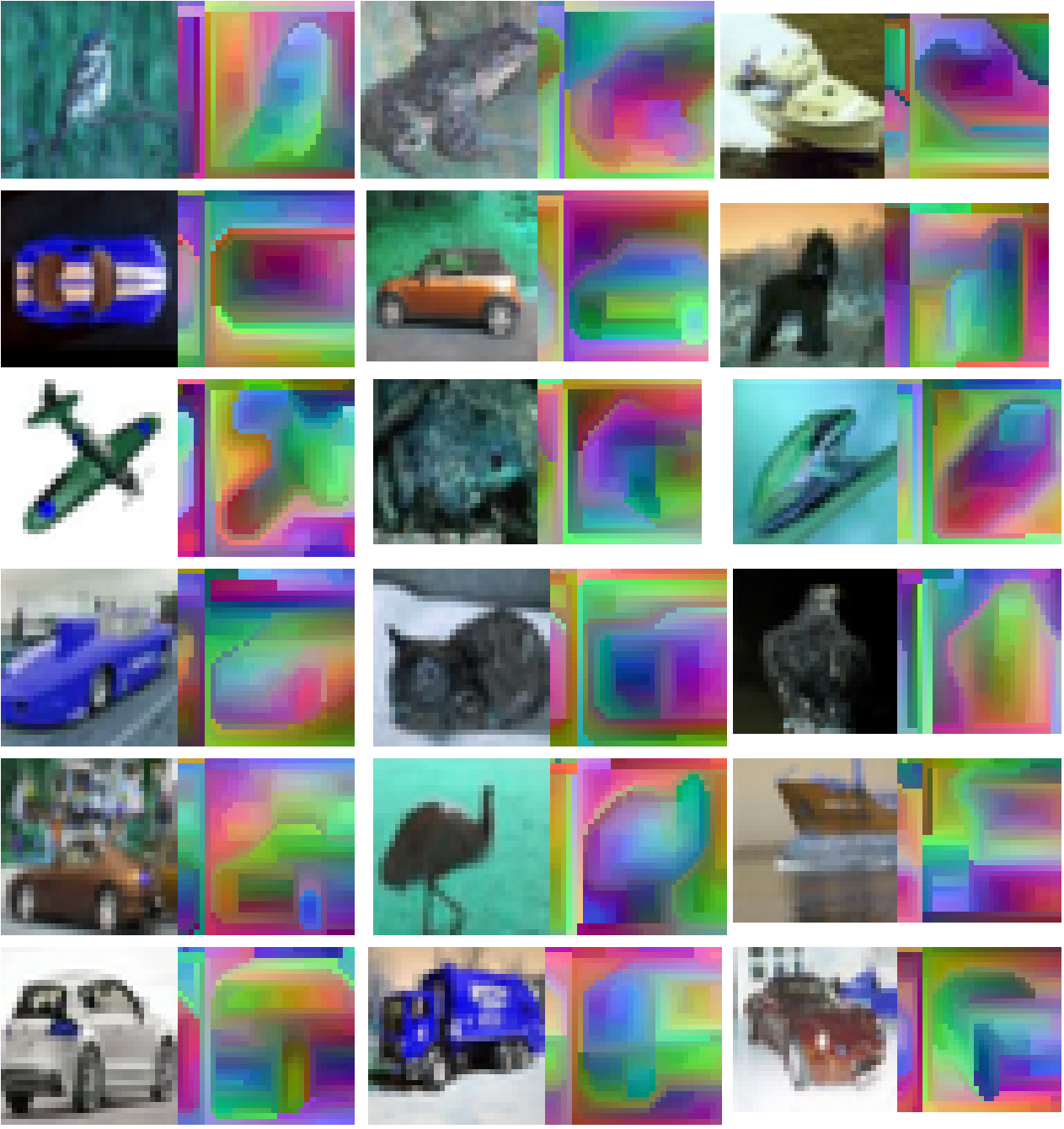}
    \caption{\textbf{Cifar 10 islands:} Notice how individual parts are clearly observed in APM features. These features are used for downstream classification. We leverage the visualization mechanism by \cite{modi2023occlusions}. These islands are \textit{not} been hand-picked.\cite{hinton2023represent} }
    \label{fig:cifar_2}
    \vspace{-1.0em}
\end{figure}

\begin{figure*}[t!]
    \centering
    \includegraphics[width=\linewidth]{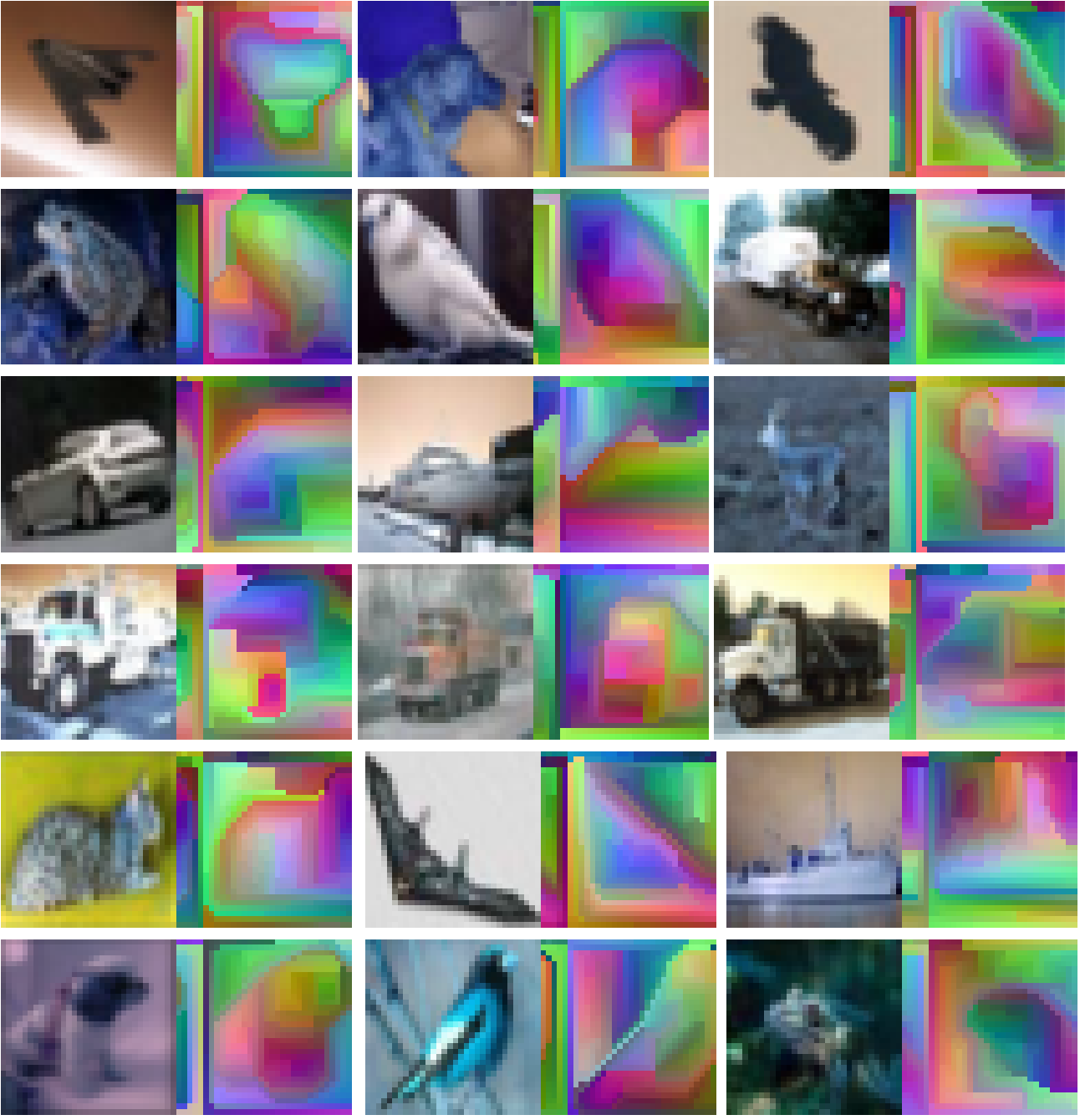}
    \caption{\textbf{Cifar 10 islands:} Notice how individual parts are clearly observed in APM features. These features are used for downstream classification.We leverage the visualization mechanism by \cite{modi2023occlusions}. These islands are \textit{not} been hand-picked.\cite{hinton2023represent} }
    \label{fig:cifar_3} 
\end{figure*}

\begin{figure*}[t!]
    \centering
    \includegraphics[width=\linewidth]{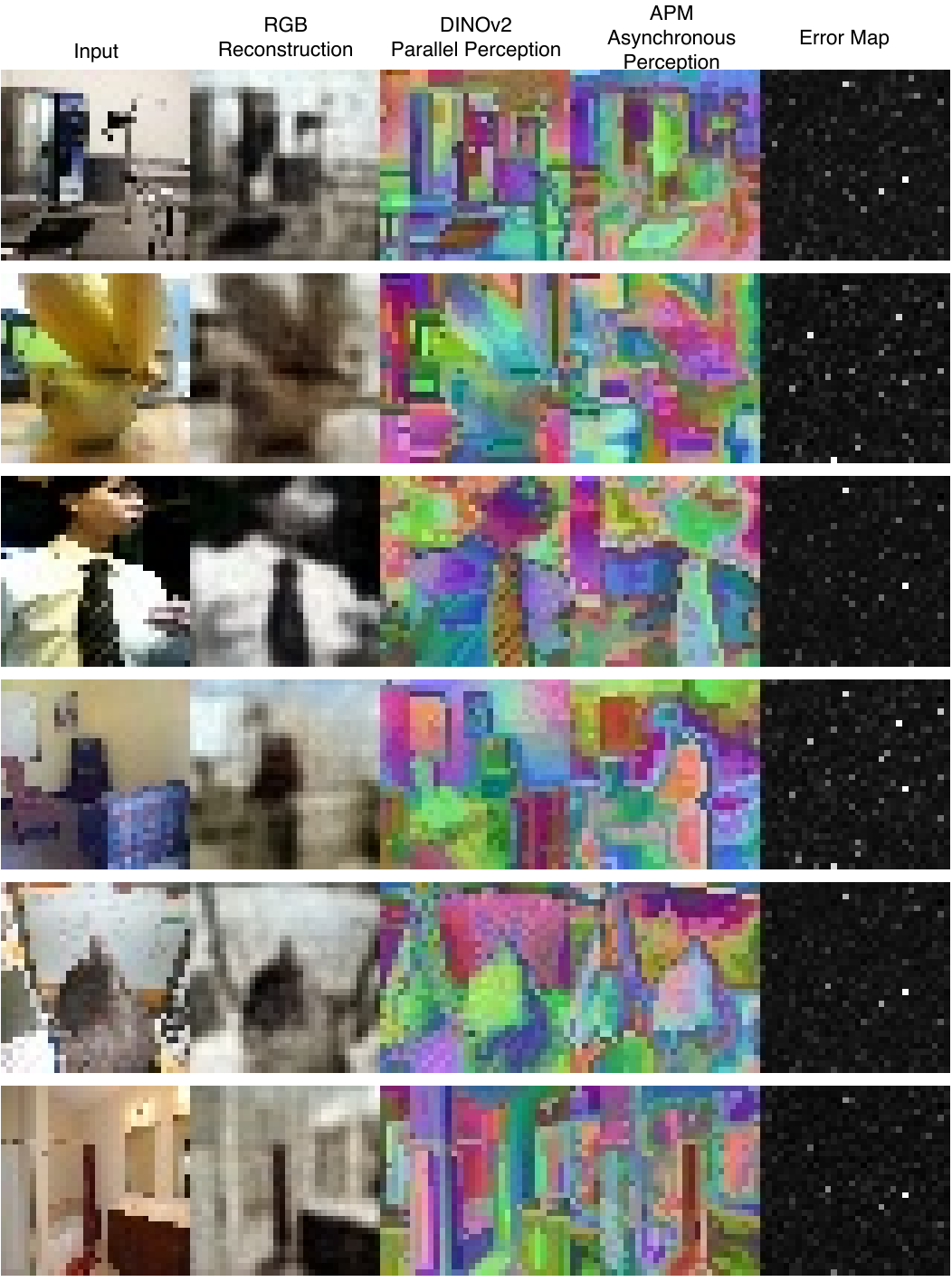}
    \caption{\textbf{Qualitative Results on COCO Val set:} Our APM is trained on COCO-train set, and islands on COCO-val set are visualized\cite{hinton2023represent}. Note that these islands are a consequence of \textit{single} feed-forward through the net and not an iterative routing mechanism as in \cite{garau2022interpretable}. DINOv2 does parallel perception: i.e. all tokens are kept in the memory. However, APM does asynchronous perception: it can predict the column vector at any location asynchronously. The error map shows the error between the grid predicted by Dinov2 and the grid predicted by APM. It is mostly black which shows APM closely approximates Dinov2 grid as well as can be memory efficient. The islands shown in this figure are \textit{not} hand-picked\cite{hinton2023represent}. }
    \label{fig:coco_2}
\end{figure*}

\begin{figure*}[t!]
    \centering
    \includegraphics[width=\linewidth]{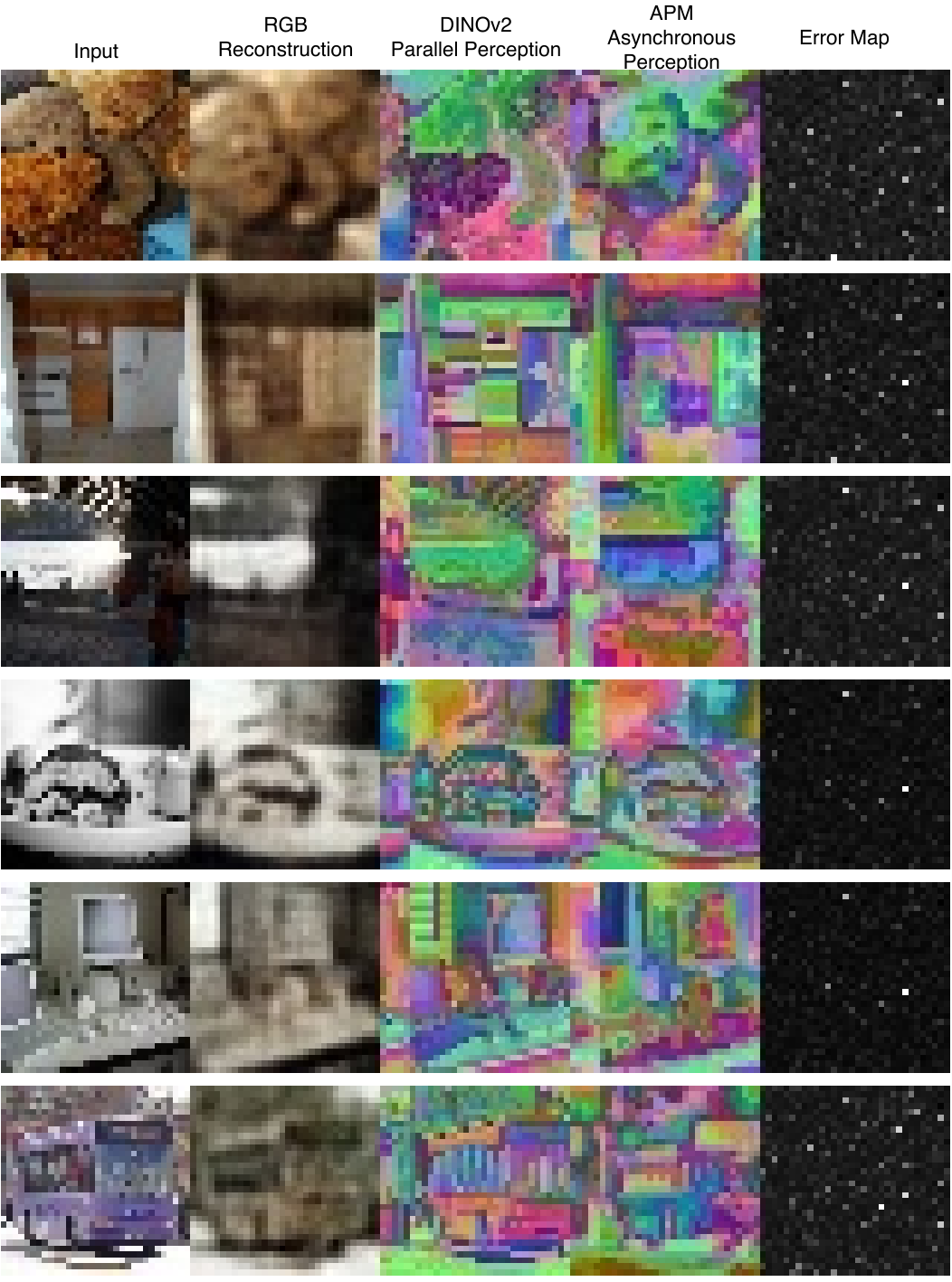}
    \caption{\textbf{Qualitative Results on COCO Val set:} Our APM is trained on COCO-train set, and islands on COCO-val set are visualized\cite{hinton2023represent}. Note that these islands are a consequence of \textit{single} feed-forward through the net and not an iterative routing mechanism as in \cite{garau2022interpretable}. DINOv2 does parallel perception: i.e. all tokens are kept in the memory. However, APM does asynchronous perception: it can predict the column vector at any location asynchronously. The error map shows the error between the grid predicted by Dinov2 and the grid predicted by APM. It is mostly black which shows APM closely approximates Dinov2 grid as well as can be memory efficient. The islands shown in this figure are \textit{not} hand-picked\cite{hinton2023represent}.}
    \label{fig:coco_1}
\end{figure*}

\begin{figure*}[t!]
    \centering
    \includegraphics[width=\linewidth]{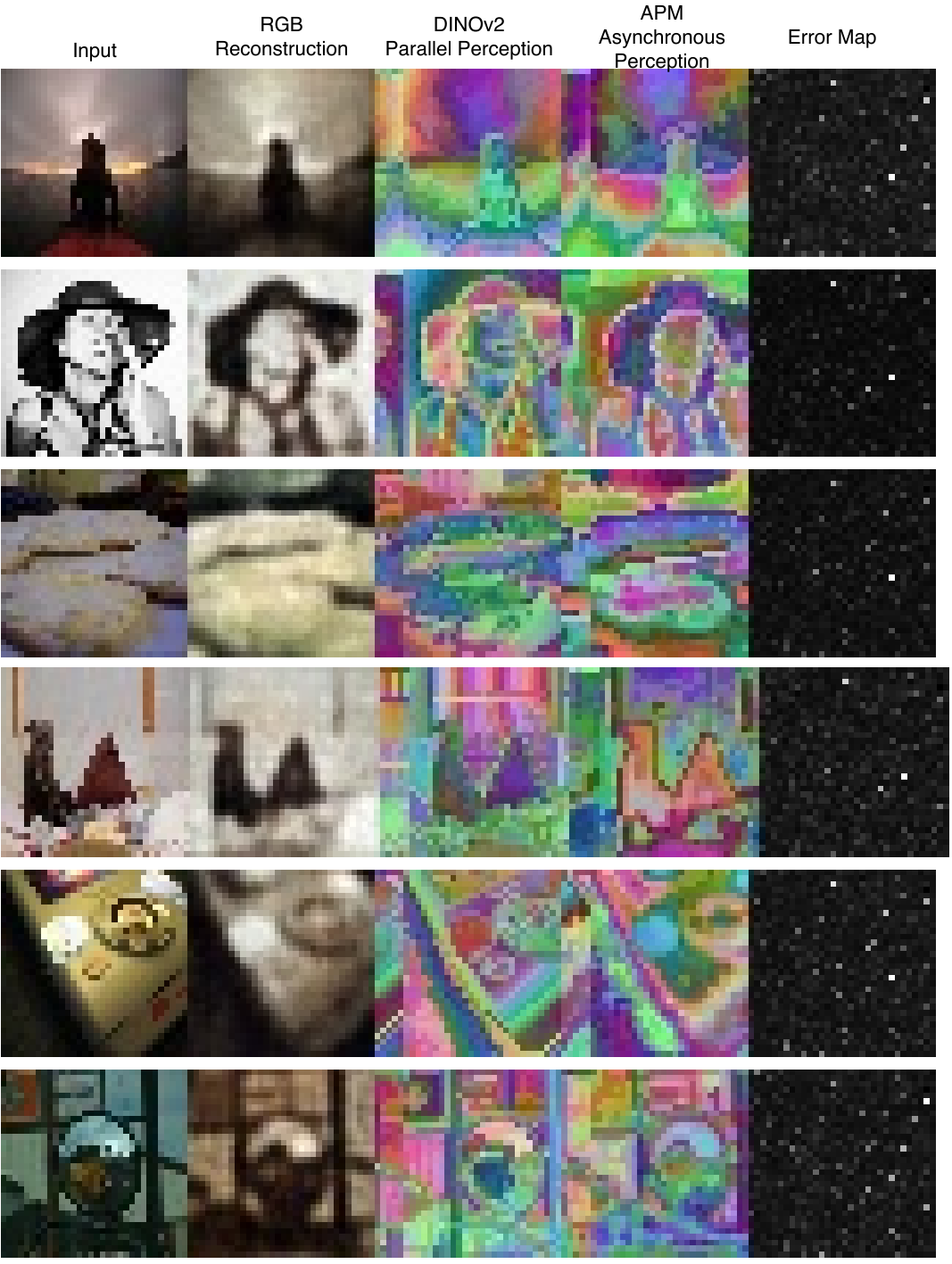}
    \caption{\textbf{Qualitative Results on COCO Val set:} Our APM is trained on COCO-train set, and islands on COCO-val set are visualized\cite{hinton2023represent}. Note that these islands are a consequence of \textit{single} feed-forward through the net and not an iterative routing mechanism as in \cite{garau2022interpretable}. DINOv2 does parallel perception: i.e. all tokens are kept in the memory. However, APM does asynchronous perception: it can predict the column vector at any location asynchronously. The error map shows the error between the grid predicted by Dinov2 and the grid predicted by APM. It is mostly black which shows APM closely approximates Dinov2 grid as well as can be memory efficient. The islands shown in this figure are \textit{not} hand-picked\cite{hinton2023represent}.}
    \label{fig:coco_4}
\end{figure*}

\newpage

\end{document}